\documentclass{ecai} 

\usepackage{amsmath,amsfonts,bm}

\def\eqref#1{equation~\ref{#1}}

\def\1{\bm{1}}

\def\rvx{{\mathbf{x}}}

\def\vy{{\bm{y}}}

\def\mA{{\bm{A}}}

\def\mI{{\bm{I}}}

\def\mX{{\bm{X}}}

\DeclareMathAlphabet{\mathsfit}{\encodingdefault}{\sfdefault}{m}{sl}
\SetMathAlphabet{\mathsfit}{bold}{\encodingdefault}{\sfdefault}{bx}{n}
\newcommand{\tens}[1]{\bm{\mathsfit{#1}}}

\def\tE{{\tens{E}}}

\def\gG{{\mathcal{G}}}

\def\gW{{\mathcal{W}}}

\def\emA{{A}}

\newcommand{\E}{\mathbb{E}}

\newcommand{\R}{\mathbb{R}}

\newcommand{\KL}{D_{\mathrm{KL}}}

\usepackage{latexsym}
\usepackage{amssymb}
\usepackage{amsmath}
\usepackage{amsthm}
\usepackage{booktabs}
\usepackage{enumitem}
\usepackage{graphicx}
\usepackage{color}
\usepackage{multirow}
\usepackage[caption=false, position=top]{subfig}
\usepackage{algorithm}
\usepackage{algpseudocode}

\usepackage{caption}
\usepackage{setspace}

\DeclareCaptionLabelSeparator{periodquad}{.\quad}

\newtheorem{definition}{Definition}

\newcommand{\BibTeX}{B\kern-.05em{\sc i\kern-.025em b}\kern-.08em\TeX}

\begin{document}


\begin{frontmatter}


\paperid{2477}


\title{IFH: a Diffusion Framework for Flexible Design\\of Graph Generative Models}


\author[A,B]{\fnms{Samuel}~\snm{Cognolato}\thanks{Corresponding Author. Email: samuel.cognolato@phd.unipd.it}}
\author[A,B]{\fnms{Alessandro}~\snm{Sperduti}}
\author[B]{\fnms{Luciano}~\snm{Serafini}} 

\address[A]{University of Padova, Padova, Italy}\address[B]{Fondazione Bruno Kessler, Trento, Italy}


\begin{abstract}
Graph generative models can be classified into two prominent families: one-shot models, which generate a graph in one go, and sequential models, which generate a graph by successive additions of nodes and edges. Ideally, between these two extreme models lies a continuous range of models that adopt different levels of sequentiality. This paper proposes a graph generative model, called Insert-Fill-Halt (IFH), that supports the specification of a sequentiality degree. IFH is based upon the theory of Denoising Diffusion Probabilistic Models (DDPM), designing a node removal process that gradually destroys a graph. An insertion process learns to reverse this removal process by inserting arcs and nodes according to the specified sequentiality degree. We evaluate the performance of IFH in terms of quality, run time, and memory, depending on different sequentiality degrees. We also show that using DiGress, a diffusion-based one-shot model, as a generative step in IFH leads to improvement to the model itself, and is competitive with the current state-of-the-art.
\end{abstract}

\end{frontmatter}



\section{Introduction}
\label{sec:intro}
Graph generative models usually fall into two categories: one-shot models, which generate the entire graph in one go, and sequential models, which iteratively extend the graph with new nodes and edges. State-of-the-art one-shot models are built upon well-established generative frameworks such as Variational Autoencoders (VAE) \citep{simonovsky2018graphvae}, Normalizing Flows \citep{zang2020moflow}, and Diffusion Models \citep{vignac2022digress}. The same techniques have been applied for developing sequential models \citep{liao2019gran, luo2021graphdf, kong2023grapharm}. This additional inductive bias may spark the idea that a better sample quality can be achieved due to regularities in the graphs \citep{jin2018junction}. However, this is not always the case, as one-shot models have entered the state-of-the-art on challenging datasets like ZINC250k \citep{irwin2012zinc}, Ego \citep{sen2008ego}, and many more. Still, they are not flexible on the size of generated graphs, which is usually sampled from the dataset empirical distribution of nodes, a pre-computed histogram of the graphs' sizes. This is not ideal in a conditional generation setup, where conditioning variables can influence the extent of the graph. In this sense, autoregressive sequential models are more flexible because they can also learn the distribution of the number of nodes. This begs the question of whether we can borrow the strength of powerful one-shot models, and enhance them with the mentioned flexibility.

\begin{figure}[t]
\centering
\includegraphics[width=0.70\columnwidth]{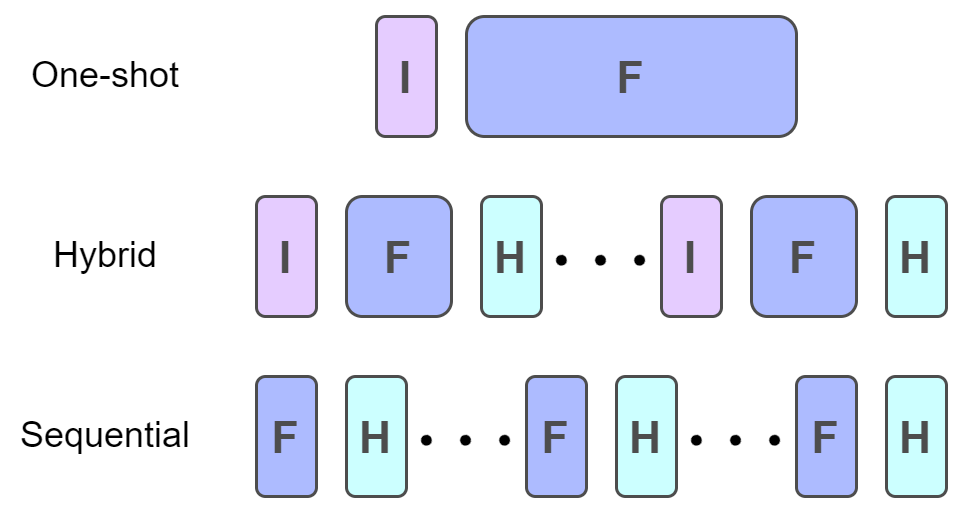}
\caption{Graph generation can be seen as a sequence of Node Insertions (I), Labels and Connections Filling (F), and Halting Choices (H). One-shot models fill graphs in 1 big step after choosing the number of nodes. 1-node sequential models add one node, fill its value, connect it with the remainder graph, and choose whether to stop or continue iterating. Our IFH framework can model these situations and the intermediate block sequential generation.}
\label{fig:spectrum}
\vspace{8pt}
\end{figure}

We answer the question by showing that, between the one-shot and sequential models, lies a continuous range of models that generate a graph by iteratively extending it more than one node at a time (see Figure~\ref{fig:spectrum}). Then, the one-shot model's most defining feature is not that of generating all nodes in one step, but the formulation of the nodes and edges sampling strategy, e.g., by sampling from the latent space with a VAE. In this paper, we propose a flexible graph framework obtained from the theory of diffusion models \cite{ho2020diffusion}, called Insert-Fill-Halt (IFH), that allows us to specify the process of choosing how many and which nodes are added at each iteration. Once a formulation on the nodes and edges sampling is fixed, IFH can reproduce the behavior of the one-shot and sequential model while simultaneously allowing all possible intermediate sequentiality levels. Specifically, IFH identifies a node removal process and an insertion process. The former gradually deletes groups of nodes, while the latter tries to add them back, together with their labels and connections. The insertion process is carried out by (1) an Insertion Model, which decides how many new nodes to generate; (2) a Filler Model, which samples the new nodes' labels and connections; (3) a Halt Model, which chooses whether to halt the insertion process. The three components can be trained on their respective tasks on partial graphs, produced by the node removal process, starting from a clean graph.

Our framework provides a mathematical foundation that can be used both to adapt sampling strategies to any sequentiality level, and design new insertion schemes by customizing the removal process. On the former, we show that any current one-shot model can act as a Filler Model inside our IFH framework. On the latter, one can schedule how many nodes to add in a step and their ordering, depending on the dataset domain and size of graphs, taking into consideration time and memory constraints. We summarize our contributions as follows:
\begin{enumerate}
    \item we propose Insert-Fill-Halt (IFH), a general graph generation framework based on Diffusion Models' theory \citep{sohl2015thermo, ho2020diffusion} that can be specialized into many old and new graph generative models;
     \item in the present work, we design two node-removal processes for IFH: naive and categorical, and use two node-orderings: random order and Breadth First Search (BFS) order. We evaluate their effectiveness in an ablation study on the QM9 molecular dataset;
     \item we show empirically that adapting Digress \citep{vignac2022digress}, a state-of-the-art one-shot graph generative model, to 1-node sequential, leads to surpassing all autoregressive baselines, and is competitive with other state-of-the-art one-shot baselines such as CDGS \citep{huang2023cdgs};
    \item we conduct a computational memory-time tradeoff survey when varying the degree of sequentiality, meaning going from one-shot to block-sequential, to 1-node sequential.
\end{enumerate}


\section{Notation and background}
\label{sec:background}
We introduce the notation we will use throughout the paper. We refer the reader to Appendix~\ref{app:defs} for a more detailed explanation. Let \mbox{$\gG=(\mathcal{V}, \mathcal{E})$} be a graph, where $\mathcal{V}=\{v_1,..,v_n\}$ is the set of vertices and $\mathcal{E}\subseteq\mathcal{V}\times\mathcal{V}$ is the set of directed arcs of $\mathcal{G}$. Let $n=|\mathcal{V}|$ and $m=|\mathcal{E}|$ be the number of nodes and edges of $\mathcal{G}$. $\mathcal{E}$ can be represented with the adjacency matrix $\mA\in \{0,1\}^{n\times n}$ where $\emA_{i,j}=1$ iff $(v_i,v_j)\in\mathcal{E}$. In the case of undirected graphs $\mA$ is a symmetric matrix, i.e., 
$\mA=\mA^\top$. If $\mathcal{G}$ is labeled, then  $\mathcal{V}$ and $\mathcal{E}$ are coupled with node features $\mX\in\R^{n\times d_{x}}$ and edge features $\tE\in\R^{n\times n\times d_{e}}$, where $d_x$ and $d_e$ are the dimensions of a single node/edge feature vector respectively. Global features are denoted as $\vy\in\R^{d_y}$.

A node $v$ can be \textit{removed} from $\gG$, meaning it is removed from $\mathcal{V}$, deleting all its connections and the labels attached to it. By removing a subset of nodes $\mathcal{V}_B\subseteq\mathcal{V}$ from a graph $\gG$ we obtain the induced subgraph $\gG_A$, with nodes $\mathcal{V}_A=\mathcal{V} \setminus \mathcal{V}_B$. The graph $\gG$ can be \textit{split} through the set of nodes $\mathcal{V}_A$ into the tuple $(\gG_A,\gG_B,\mathcal{E}_{AB},\mathcal{E}_{BA})$, where $\gG_A$ and $\gG_B$ are the subgraphs induced by $\mathcal{V}_A$ and $\mathcal{V}_B$ respectively, and $\mathcal{E}_{AB},\mathcal{E}_{BA}$ are the edges connecting them in one direction and the other. The inverse operation is a \textit{merge}, which merges the two induced subgraphs and the edges back into $\gG$. One of the main mathematical objects for this paper is the \textit{forward graph removal sequence} $\gG_{0:T}^{\rightarrow}=(\gG_t)_{t=0}^T$ for graph $\gG$, which is any sequence such that $\gG_0=\gG$, $\gG_T$ is the empty graph $\varnothing$, and $\gG_t$ is an induced subgraph of $\gG_{t-1}$ for all $t=1,\dots,T$. By reversing the order of $\gG_{0:T}^{\rightarrow}$ we obtain the \textit{reversed graph removal sequence} $\gG_{0:T}^{\leftarrow}$, which will be useful to define the generative process from $\gG_0=\varnothing$ to $\gG_T=\gG$ as the resulting graph. We denote $\mathcal{F}(\gG,T)$ and $\mathcal{R}(\gG,T)$ as the sets of all forward and reversed removal sequences of $\gG$ of length $T$, respectively.

Additionally, we define the halting process, borrowing the notation from \cite{banino2021pondernet}. $\Lambda_t$ is a Markov chain with two states: \textit{continue}, \textit{halt}. The chain starts in the \textit{continue} state, and proceeds at each step $t$ with probability $\lambda(t)$ of being absorbed into the \textit{halt} state. Once there, the process is stuck forever in the \textit{halt} state.


\section{Related works}
\label{sec:rel_work}

\subsection{Graph generation}
\label{sec:rel_work/graph_gen}

Given a set of graph data points with unknown distribution $p_\text{data}(\gG)$, likelihood maximization methods aim to learn the parameters $\theta$ of a model $p_\theta(\gG)$ to approximate the true distribution $p_\text{data}(\gG)$. In the context of deep graph generation \citep{guo2022systematic}, $p_\theta(\gG)$ has been modeled as one-shot and sequential models.

\paragraph{One-shot models}
One-shot models employ a decoder network that maps a latent vector $z$ to the resulting graph $\gG$. The latent vector is usually sampled from a tractable distribution (such as a Normal distribution), and the number of nodes is either fixed, sampled from the frequencies of nodes in the dataset, or predicted from the latent code $z$. In general, one-shot models have the form:
\begin{equation}
p_{\theta,\phi}(\gG) = p_\theta(\gG|n)p_\phi(n).   \label{eq:one_shot}
\end{equation}
When $p_\theta(\gG|n)$ is implemented by a neural network architecture equivariant to node permutations, no node orderings are needed.

For one-shot generation, the classic generative paradigms are applied: VAE with GraphVAE \citep{simonovsky2018graphvae}, GAN with MolGAN \citep{de2018molgan} and SPECTRE \citep{martinkus2022spectre}, Normalizing Flows with MoFlow \citep{zang2020moflow}, diffusion with EDP-GNN \citep{niu2020edp_gnn}, discrete diffusion with DiGress \citep{vignac2022digress}, energy-based models with GraphEBM \citep{liu2021graphebm}, Stochastic Differential Equations (SDE) with GDSS \citep{jo2022gdss} and CDGS \citep{huang2023cdgs}.

\paragraph{Sequential models}
Sequential models frame graph generation as forming a sequence $\gG_{0:T}^{\leftarrow}=(\gG_0,\dots,\gG_t,\dots,\gG_T)$ of increasingly bigger graphs, where $\gG_0$ is usually an empty graph, $\gG_T$ is the generation result, and the transition from $\gG_{t-1}$ to $\gG_t$ introduces new nodes and edges. In the case of node-sequence generation, transitions always append exactly one node and the edges from that node to $\gG_{t-1}$. In motif-sequence generation, blocks of nodes are inserted, together with new rows of the adjacency matrix. For the remainder of the paper, we will denote as \textit{1-node sequential} the models based on a node-sequence generation, \textit{block-sequential} for motif-based models, and \textit{autoregressive models} for addressing both. Given a halting criteria $\lambda_{\nu}(\gG_t, t)$ (see Section~\ref{sec:background}) based on current graph $\gG_t$, the model distribution for a 1-node sequential model is of the form:
\begin{equation}
\begin{split}
    p_{\theta,\nu}(\gG)= & \sum_{\gG_{0:n}^{\leftarrow} \in \mathcal{R}(\gG,n)}\lambda_\nu(\gG_n,n)p_\theta(\gG_{n} | \gG_{n-1}) \\
    & \cdot p_\theta(\gG_{0}) \prod_{t=1}^{n-1} (1-\lambda_\nu(\gG_t, t)) p_\theta(\gG_{t} | \gG_{t-1}).
    \label{eq:sequential}
    \end{split}
\end{equation}
An essential ingredient in training autoregressive models is the node ordering, i.e., assigning a permutation $\pi$ to the $n$ nodes in $\gG$ to build the sequence $\gG_{0:T}^{\leftarrow}$. With random ordering, the model has to explore $n!$ permutations. In contrast, canonical orderings such as Breadth First Search (BFS) \citep{you2018graphrnn}, Depth First Search (DFS), and many others \citep{liao2019gran, chen2021order}, decrease the size of the search space by introducing inductive biases, empirically increasing sample quality in most instances.

\subsection{Diffusion models}
\label{sec:rel_work/diff_models}

We briefly introduce the Diffusion Model \citep{sohl2015thermo, ho2020diffusion}, on which we build IFH. Let $\rvx_0$ be a data point sampled from an unknown distribution $q(\rvx_0)$. Denoising diffusion models are latent variable models with two components: (1) a diffusion process gradually corrupts $\rvx_0$ in $T$ steps with Markov transitions $q(\rvx_t|\rvx_{t-1})$ until $\rvx_T$ has some tractable distribution $p_\theta(\rvx_T)$; (2) a learned reverse Markov process with transition $p_\theta(\rvx_{t-1}|\rvx_t)$ denoises $\rvx_T$ back to the original data distribution $q(\rvx_0)$. The trajectories formed by the two processes are:
\begin{align}
    q(\rvx_{1:T}|\rvx_{0}) &= \prod_{t=1}^T q(\rvx_t|\rvx_{t-1}), & \text{Forward} \\
    p_\theta(\rvx_{0:T}) &= p_\theta(\rvx_T)\prod_{t=1}^T p_\theta(\rvx_{t-1}|\rvx_t) & \text{Reverse} \label{eq:for_rev_proc}
\end{align}
For $T\to +\infty$, the forward and reverse transitions share the same functional form \citep{1949bsms.conf..403F}, and choosing $q(\rvx_T|\rvx_0)=q(\rvx_T)$ allows in fact to easily sample $\rvx_T$. The first successful attempt with diffusion models defined the transitions as $q(\rvx_t|\rvx_{t-1})=\mathcal{N} ( \rvx_t ; \sqrt{1-\beta_t} \rvx_{t-1}, \beta_t \mI)$ \citep{ho2020diffusion} where $\beta_t$ is a variance schedule. Later, diffusion models were adapted for discrete spaces \citep{austin2021discdiff}, introducing concepts like uniform transitions, used in DiGress \citep{vignac2022digress} with node and edge labels, and absorbing states diffusion, adopted in GraphARM \citep{kong2023grapharm} for masking nodes.

The distribution $p_\theta(\rvx_0)$ can be made to fit the data distribution $q(\rvx_0)$ by minimizing the variational upper bound:
\begin{align}
    L_\text{vub}=&\E_{\rvx_0\sim q(\rvx_0)}\bigg[\underbrace{\KL \big(q(\rvx_T|\rvx_0) \Vert p(\rvx_{T})\big)}_{L_T}+ \nonumber\\
	&+\sum\limits_{t=2}^{T}\underbrace{\KL \big(q(\rvx_{t-1}|\rvx_{t},\rvx_{0}) \Vert p_\theta(\rvx_{t-1}|\rvx_t)\big)}_{L_{t-1}}+ \nonumber\\
	&\underbrace{-\E_{\rvx_1\sim q(\rvx_1|\rvx_0)}\left[\log p_\theta(\rvx_{0}|\rvx_1)\right]}_{L_0}
	\bigg].
 \label{eq:diff_vub}
\end{align}
A necessary property to make diffusion models feasible to train is for $q(\rvx_t|\rvx_0)$ and $q(\rvx_{t-1}|\rvx_t,\rvx_0)$ to have a closed form formula, to respectively (1) efficiently sample many time steps in parallel and (2) compute the KL divergences.



\section{Removing nodes as a graph noise process}
\label{sec:rem_proc}

\begin{figure*}[t]
\begin{center}
\includegraphics[width=1.\textwidth]{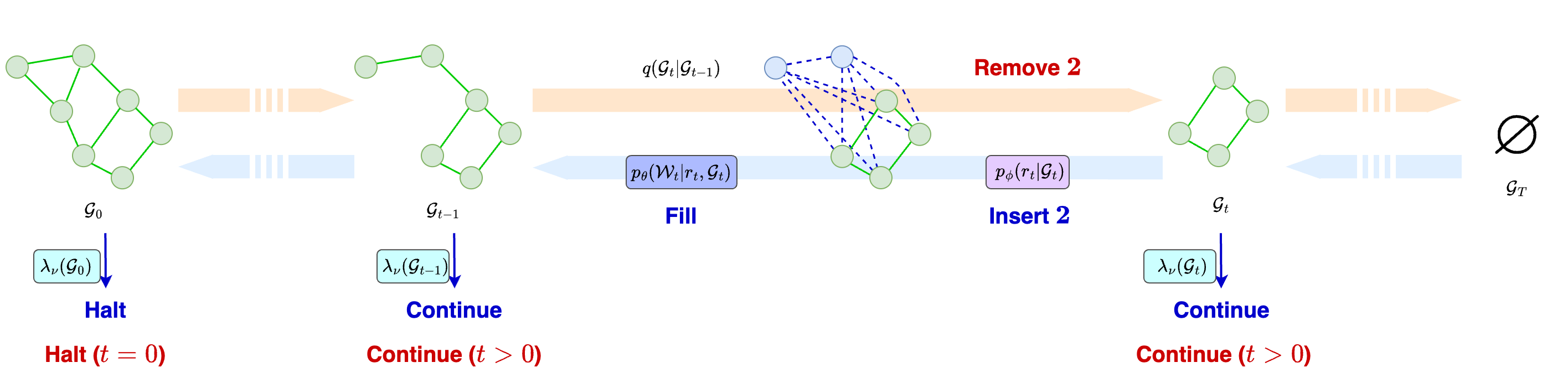}
\end{center}
\vspace*{-3mm}
\caption{Our Insert-Fill-Halt model. During training, a graph is corrupted (left to right) by iteratively removing nodes until the empty graph $\varnothing$ is left. At each step, the insertion (violet), filler (blue), and halt (cyan) models have to predict how many nodes were removed, what content they had, and whether the graph is terminal, respectively (right to left).}
\label{fig:ifh}
\end{figure*}

In this work, we frame the process of removing nodes from a graph $\gG$ as an absorbing state diffusion process \citep{austin2021discdiff}, gradually corrupting $\gG$ until it collapses to the empty graph $\varnothing$. Differently from the absorbing diffusion of GraphARM \citep{kong2023grapharm}, we do not limit the process to the choice of one node per step, but we include both the node ordering and number of nodes removed in the transitions.

Formally, given a graph data point $\gG$, the removal process generates a removal sequence $\gG_{0:T}^{\rightarrow}$ with $\gG_0=\gG$ and $\gG_T$ being the empty graph. We define the Markov removal transition $q(\gG_t|\gG_{t-1})$ as the probability of sampling a set of nodes $\mathcal{V}_t\subseteq\mathcal{V}_{t-1}$, and computing the induced subgraph $\gG_t$ from $\gG_{t-1}$ by $\mathcal{V}_t$. Following from eq.~(\ref{eq:for_rev_proc}), the forward process is defined as:
\begin{equation}
q(\gG_{1:T}^{\rightarrow}|\gG_0)=\prod_{t=1}^T q(\gG_t|\gG_{t-1}).
    \label{eq:remv_proc}
\end{equation}

Now we show the key insight that, because the number of nodes \mbox{$n_t=|\mathcal{V}_{t}|$} is a known property of $\gG_t$, the removal transition can be factorized into two components:
\begin{align}
    q(\gG_t|\gG_{t-1}) = q(\gG_t,n_t|\gG_{t-1})= 
    q(\gG_t|n_t,\gG_{t-1})q(n_t|\gG_{t-1}),
    \label{eq:broken_remv}
\end{align}
where $q(n_t|\gG_{t-1})$ is the probability that $\mathcal{V}_t$ will have exactly $n_t$ nodes, and, fixed this number, $q(\gG_t|n_t,\gG_{t-1})$ is the probability of choosing the nodes in $\mathcal{V}_t$ from $\mathcal{V}_{t-1}$. In simpler words, $q(n_t|\gG_{t-1})$ tells \textit{how many} nodes to keep alive, and once this fact is known, $q(\gG_t|n_t,\gG_{t-1})$ select \textit{which} nodes. In some special cases of the removal process, we will show that the number of nodes $n_{t-1}$ is enough information to sample $n_t$, i.e., $q(n_t|\gG_{t-1})=q(n_t|n_{t-1})$.

\subsection{Parameterizing the reverse of the removal process}
\label{sec:rem_proc/paramet}

Again, following the theory of diffusion models (Section~\ref{sec:rel_work/diff_models}), we introduce the insertion process, which learns to regenerate the graphs corrupted by the removal process. Define $p_{\theta,\phi}(\gG_{t-1}|\gG_t)$ as the Markov insertion transition which, given a partial graph $\gG_t=\gG_{t,A}$, samples a new subgraph $\gG_{t,B}=(\mathcal{V}_{t,B}, \mathcal{E}_{t,B})$ with $r_t=n_{t-1}-n_t$, together with edges $\mathcal{E}_{t,AB},\mathcal{E}_{t,BA}$ to connect the two graphs. Then, through a merge operation (as explained in Section~\ref{sec:background}), graph $\gG_{t-1}$ is composed. The process reversing eq.~(\ref{eq:remv_proc}) is defined as:
\begin{equation}
    p_{\theta,\phi}(\gG_{0:T}^{\rightarrow})=\prod_{t=1}^T p_{\theta,\phi}(\gG_{t-1}|\gG_t),
    \label{eq:reins_proc}
\end{equation}
where we omitted the $p_{\theta,\phi}(\gG_T)$ term, as all the probability mass is already placed on the empty graph $\varnothing$. Again, we can factorize the transition into two components 
\begin{equation}
    p_{\theta,\phi}(\gG_{t-1}|\gG_t) = p_{\theta,\phi}(\gG_{t-1},r_t|\gG_t)
    =p_{\theta}(\gG_{t-1}|r_t,\gG_t)p_{\phi}(r_t|\gG_t),
    \label{eq:broken_reins}
\end{equation}
where we call $p_{\phi}(r_t|\gG_t)$ the \textit{insertion model}, with parameters $\phi$, and $p_{\theta}(\gG_{t-1}|r_t,\gG_t)$ the \textit{filler model}, with parameters $\theta$. The role of each is respectively: (1) given a partial subgraph $\gG_t$ decide how many nodes $r_t$ to add, (2) known this number and $\gG_t$, generate the content of the new nodes and respective edges, and how to connect them to $\gG_t$. Expanding $p_{\theta,\phi}(\gG_{t-1},r_t|\gG_t)$, we have:
\begin{align}
    p_{\theta,\phi}&(\gG_{t,A},\gG_{t,B},\mathcal{E}_{t,AB},\mathcal{E}_{t,BA}, r_t|\gG_{t,A})= \nonumber \\
    &=p_{\theta}(\gG_{t,B},\mathcal{E}_{t,AB},\mathcal{E}_{t,BA}|r_t,\gG_{t,A})p_{\phi}(r_t|\gG_{t,A})= \nonumber \\
    &=p_{\theta}(\gW_t|r_t,\gG_t)p_{\phi}(r_t|\gG_t)
    \label{eq:broken_reins_2},
\end{align}
where we packed the tuple $\gW_t=(\gG_{t,B},\mathcal{E}_{t,AB},\mathcal{E}_{t,BA})$ for brevity. In Appendix~\ref{app:proofs}, we show that, similarly to eq.~(\ref{eq:diff_vub}), when $q(r_t|\gG_t,\gG_0)$ and $q(\gW_t|r_t,\gG_t,\gG_0)$ can be expressed in closed form they can be estimated by minimizing the variational upper bound:
\begin{align}
    L_\text{vub}=&\E_{\gG_0\sim q(\gG_0)}\Bigg[
	\sum\limits_{t=2}^{T}\KL \big(q(r_t|\gG_t,\gG_0) \Vert p_{\phi}(r_t|\gG_t)\big) + \nonumber\\
    &-\E_{\gG_1\sim q(\gG_1|\gG_0)}\left[\log p_\phi(r_{1}|\gG_1)\right]+ \nonumber\\
	&+\sum\limits_{t=2}^{T}\KL \big(q(\gW_t|r_t,\gG_t,\gG_0) \Vert p_{\theta}(\gW_t|r_t,\gG_t)\big)+ \nonumber\\
    &-\E_{\gG_1\sim q(\gG_1|\gG_0)}\left[\log p_\theta(\gW_1|r_1,\gG_1)\right]
	\Bigg].
    \label{eq:remv_vub}
\end{align}
The KL divergence of the filler model term can be replaced by the negative log-likelihood, at the price of increasing the upper bound. On the other hand, this allows to train $p_{\theta}(\gW_t|r_t,\gG_t)$ through any likelihood maximization method, such as VAE, Normalizing Flow, and Diffusion. In particular, noticing the resemblance with eq.~\ref{eq:one_shot}, the filler model can be any likelihood-based one-shot model, although adapted to be conditioned on an external graph $\gG_t$, and able to generate the interconnections, which we explain in Section~\ref{sec:spect/adapting}.
\vspace{-4pt}

\subsection{Choosing the removal process}
\label{sec:rem_proc/choose}

The design of $q(\gG_t|\gG_{t-1})$ influences the graph generative model $p_{\theta,\phi}(\gG_{t-1}|\gG_{t})$ in several ways. We start by proposing a naive \textit{coin flip} approach for removing nodes, moving afterwards to a more effective way of choosing the number of nodes to remove, and how to incorporate node ordering. All proofs can be found in Appendix~\ref{app:proofs}.

\paragraph{Naive/binomial (Appendix~\ref{app:remv_proc/naive})}
The simplest way to remove nodes from a graph $\gG_{t-1}$ is to assign a Bernoulli random variable with probability $q_t$ to each node. All nodes with a positive outcome are removed. It follows that, given $n_{t-1}$ nodes at step $t-1$, the count of survived nodes $n_t$ is distributed as a Binomial $B(n_t; n_{t-1}, 1-q_t)$. Iterating this process for $t$ steps from graph $\gG_0$, we still get that $n_t|n_0$ is distributed as a Binomial, where the probability of being alive at step $t$ is the product of being alive at all steps. Finally, the posterior $q(\gG_t|n_t,\gG_{t-1})$, needed for computing the loss, is again distributed as a Binomial on the removed nodes $\Delta n_t=n_{0}-n_{t}$.

\paragraph{Categorical (Appendix~\ref{app:remv_proc/cat})}
One drawback of the binomial removal is that, in principle, any block size can be sampled. This can be a problem when batching multiple examples (see Section~\ref{sec:spect/complexity}), and leads to a considerable variability in block size in the training examples. To control the size of blocks generated while limiting the options available to the model, we developed a categorical removal process where the Insertion Model can choose from a predefined set of options. We based our formulation on the change-making problem \citep{wright1975change}, interpreting the number of nodes as the amount to be made using a set of coin denominations $D=\{d_1,..,d_c\}$. Then, a removal transition is defined as a categorial distribution over $D$, where each denomination's probability is its normalized count to reach $n$ with the lowest number of coins. We find that categorical removals also admit a closed form for $n_t|n_0$, distributed as a multivariate hypergeometric, and $\Delta n_t$, distributed as a mirrored version of $n_t$.

\paragraph{Node ordering (Appendix~\ref{app:remv_proc/ordering})}
Until now, we assumed nodes were removed uniformly in all possible permutations. This doesn't need to be the case, as the whole removal process can be conditioned on a particular node ordering $\pi$. The transitions will then obey $
    q(\gG_t|\gG_{t-1},\pi) = q(\gG_t|n_t,\gG_{t-1},\pi)q(n_t|\gG_{t-1},\pi)
    =q(n_t|\gG_{t-1},\pi).
$

\paragraph{Halting process}
Due to the arbitrary size of graphs, one needs to know when to stop sampling. One possibility is to stop after a fixed number of steps $T$, or when some property of the removal process is met (e.g., in the binomial process). Learning a halting process (see Section~\ref{sec:background}) can be a solution when this is not possible. A halting model $\lambda_\nu(\gG_t,t)$ can be trained in a binary classification setup by setting the halting ground truth signal to $1$ for the actual data graph $\gG$, and $0$ for all its induced subgraphs.


\section{Uncovering the spectrum of sequentiality}
\label{sec:spect}

One-shot and sequential graph generative models (Section~\ref{sec:rel_work/graph_gen}) are seen as two different families of graph generative models. Here we show that these are actually the two extremes of a spectrum, captured by our Insert-Fill-Halt (IFH) framework (Figure~\ref{fig:ifh}). First of all, let's consider the reversed removal sequence (see Section~\ref{sec:background}) $\gG_{0:T}^{\leftarrow}=(\gG_t)_{t=0}^{T}=(\varnothing,\dots,\gG)$. The three modules are: (1) a \textit{Node Insertion Module} $p_\phi(r_{t-1}|\gG_{t-1})$, deciding how many nodes are going to be inserted in $\gG_{t-1}$; (2) a \textit{Filler Module} $p_\theta(\gW_{t-1}|r_{t-1},\gG_{t-1})$, filling the new $r_{t-1}$ nodes and edges $\gW_{t-1}$, merging them with the existing graph $\gG_{t-1}$ to get $\gG_{t}$; (3) a \textit{Halting Module} $\lambda_\nu(\gG_{t})$, determining, through some halting criteria, whether to stop the generative process at $t$ or to continue. The overall model distribution is:
\begin{align}
    p&_{\theta,\phi,\nu}(\gG) =
    \sum_{T=1}^{\infty } \sum_{\gG_{0:T}^{\leftarrow} \in \mathcal{R}(\gG,T)} p_\theta(\gG_{0}) \nonumber\\
    &\underbrace{\lambda_\nu(G_{T})}_{\text{halt at last step}} p_\theta(\gW_{T-1}|r_{T-1},\gG_{T-1})p_\phi(r_{T-1}|\gG_{T-1}) \nonumber\\
    &\prod_{t=1}^{T-1}\underbrace{(1-\lambda_\nu(\gG_{t}))}_{\text{do not halt}} \underbrace{p_\theta(\gW_{t-1}|r_{t-1},\gG_{t-1})}_{\text{fill}} \underbrace{p_\phi(r_{t-1}|\gG_{t-1})}_{\text{insert}}.
    \label{eq:ifh}
\end{align}

\subsection{Specializing to one-shot and sequential models}
\label{sec:spect/special}

\paragraph{One-shot}
One-shot models (eq.~(\ref{eq:one_shot})) 
are 1-step instances of our IFH model with the insertion module set to be a sampler of the total number of nodes, i.e., $p_\phi(r_0|\varnothing)=p_\phi(n_1)=p_\phi(n)$. The filler model is the actual one-shot model, sampling all nodes in one go. The halting model always stops after 1 step.

\paragraph{Sequential}
Sequential models (eq.~(\ref{eq:sequential})) 
are $n$-step instances of our IFH model, with the insertion module always choosing 1 as the nodes to insert. The filler model samples a new node and links it with graph $\gG_{t-1}$ to compose $\gG_t$. The halting model is dependent on the architecture: in \cite{you2018graphrnn}, an End-Of-Sequence (EOS) token is sampled to end generation; in \cite{shi2020graphaf} it is not clear, but we assume they fix the number of nodes at the start; in \cite{luo2021graphdf} generation stops when a limit on $n$ is reached, or if the model does not link the newly generated node to the previous subgraph; \cite{han2023graph_halt} trains a neural network to predict the halting signal from the adjacency matrix.

\paragraph{Differences from GRAN}
GRAN \citep{liao2019gran} is a block-sequential model that upon a superficial analysis could be considered similar to our IFH. A key difference with IFH is that GRAN actually generates blocks until reaching a maximum number of nodes. This number is fixed, computed as the biggest multiple of the block size nearest to the maximum number of nodes of the dataset. Then, the actual number of nodes $n$ is sampled from the empirical distribution of nodes, and the subgraph with the first $n$ nodes is extracted. In a sense, GRAN acts as a one-shot model regarding the number of nodes, although nodes are filled in an autoregressive way. This means that GRAN does not learn the nodes distribution, and does not adapt the number of generation steps to size, implying that many sequential models cannot be built from GRAN.

\subsection{Training}

Given an example graph $\gG$, training is performed over the entire removal sequence $\gG_{0:T}^{\rightarrow}$. The models parameters $\phi,\theta,\nu$ can be optimized by minimizing the loss function:
\begin{equation}
    \mathcal{L}(\phi,\theta,\nu) = L_\text{vub}(\phi,\theta) + L_\text{halt}(\nu),
\end{equation}
with $L_\text{vub}(\phi,\theta)$ defined as in eq.~(\ref{eq:remv_vub}), and the halting loss $L_\text{halt}(\nu)$ is an optional binary classification loss as described in Section~\ref{sec:rem_proc/choose}. The full training algorithm is provided in Algorithm~\ref{alg:training}. Notice that, apart from the removal sequence sampling, every iteration of the while loop can be computed in parallel on GPU, by batching all the obtained graphs.

\begin{algorithm}[H]
\caption{Training}\label{alg:training}
\begin{algorithmic}[1]
\Repeat
   \State $\gG_0\sim q(\gG)$
   \While {$\gG_{t-1} \neq \varnothing$}
       \State $\gG_{t} \sim q(\gG_{t}|\gG_{t-1})$ \Comment{remove nodes}
       \State $r_t \gets n_{t-1} - n_t$ \Comment{get true number of nodes}
       \State $\gW_{t} \gets \operatorname{split}(\gG_{t-1}, \gG_t)$ \Comment{get true nodes and edges}
       \State $h_{t} \gets \delta(t-1)$ \Comment{get true halting signal}
       \State $L_{\text{ins},t}(\phi) \gets \KL \big(q(r_t|\gG_t,\gG_0) \Vert p_{\phi}(r_t|\gG_t)\big)$
       \State $L_{\text{fill},t}(\theta) \gets \KL \big(q(\gW_t|r_t,\gG_t,\gG_0) \Vert p_{\theta}(\gW_t|r_t,\gG_t)\big)$
       \State $L_{\text{halt},t}(\nu) \gets \mathcal{L}_{\text{halt}}(h_{t}, \lambda_\nu(\gG_{t-1}))$
   \EndWhile
   \State Perform gradient descent step on \vspace{-6pt}
   \begin{equation*}
   \vspace{-6pt}
   \frac{1}{T}\sum_{t=1}^T (L_{\text{ins},t}(\phi) + L_{\text{fill},t}(\theta) + L_{\text{halt},t}(\nu))
   \end{equation*}
\Until {converged}
\end{algorithmic}
\end{algorithm}

\subsection{Sampling}
The sampling process is a sequence of Insert, Fill, Halt operations, which is terminated by a positive halting signaling (Algorithm~\ref{alg:sampling}).

\begin{algorithm}[H]
\caption{Sampling}\label{alg:sampling}
\begin{algorithmic}[1]
    \State $\gG_0 \gets \varnothing$ \Comment{start from the empty graph}
    \Repeat
        \State $r_{t} \sim p_{\phi}(r_{t}|\gG_{t})$ \Comment{sample how many nodes to add}
        \State $\gW_{t} \sim p_{\theta}(\gW_{t}|r_{t},\gG_{t})$ \Comment{sample new nodes and edges}
        \State $\gG_{t+1} \gets \operatorname{merge}(\gG_t, \gW_t)$
        \State $h_t \sim \lambda_\nu (\gG_{t+1})$ \Comment{sample halting signal}
    \Until {$h_t=1$}
    \State return $\gG_T$
\end{algorithmic}
\end{algorithm}

\subsection{Adapting one-shot models to sequential}
\label{sec:spect/adapting}

In Section~\ref{sec:spect/special}, we showed how one-shot models are 1-step IFH models, and our parametrization in Section~\ref{sec:rem_proc/paramet} allows the use of any one-shot model inside a multi-step instance. Usually, one-shot models operate by sampling $n$ nodes and the $n\times n$ adjacency matrix. For this reason, they need to be adapted to generate the edges linking the new nodes with a previous subgraph, and to condition the former on the latter. Consider the already generated subgraph $\gG_{t-1}$, and denote $\gW_{t-1}$ as the new subgraph of size $r_{t-1}$ and the inter-connections with $\gG_{t-1}$ to be sampled. We propose the following adaptation to the $T$-step setup for undirected graphs: (1) encode the $n_{t-1}$ nodes of graph $\gG_{t-1}$ into vector representations through a Graph Neural Network such as GraphConv \citep{morris2019graphconv} or RGCN \citep{schlichtkrull2018rgcn} for labeled data; (2) generate the new $r_{t-1}$ nodes and a rectangular adjacency matrix with size $r_{t-1}\times n_s$ using the encoded node vectors, where $n_t=r_{t-1}+n_{t-1}$; (3) merge $\gG_{t-1}$ and $\gW_{t-1}$ into $\gG_t$ by concatenating nodes and the adjacency matrix. Summarizing, the strategy entails adding $r_{t-1}$ new rows to the previous adjacency matrix, without materializing it. We motivate this choice in the following section.

\subsection{Complexity considerations}
\label{sec:spect/complexity}

One-shot models generating adjacency matrices have a quadratic dependency on the number of nodes for both time and memory. However, they are very fast to train and sample from using parallelizable computing architectures such as GPUs. It is not the case for autoregressive models where, due to their iterative nature, they cannot fully benefit from parallelization \citep{liao2019gran}. Still, these do not need to generate the whole adjacency matrix in one go, and can more efficiently store the already-generated graph representation, e.g., converting to a sparse edge list (as implemented in Torch Geometric \cite{torch_geom}).
Another factor affecting memory and time is \textit{batching}, that is, generating or training on many graphs simultaneously, stacking their features in tensors. For dense representations, like adjacency matrices, the size of the resulting batched tensor is always the biggest of the batch, and the rest are padded with zeroes. This implies that memory consumption depends on the maximum size of generated blocks, so one-shot models fall on the most expensive side. This is true both when training and sampling. Still, when parallelizing training of autoregressive models on all steps, the price is paid by replicating the same example many times, just with masked nodes. We show empirically in Section~\ref{sec:experim} that these considerations are confirmed in reality.


\section{Experiments}
\label{sec:experim}
We experimentally evaluate how changing the formulation of the removal process changes sample quality, time, and memory consumption. In GRAN \cite{liao2019gran} a sample quality/time trade-off analysis on a grid graphs dataset was already performed, changing the fixed block size, stride, and node ordering. We extend this analysis to many more molecular and generic graph datasets, evaluating different degrees of sequentiality, i.e., scheduled sizes of blocks.

To showcase our framework\footnote{Our code can be found at \url{https://github.com/CognacS/ifh-model-graphgen}}, we adapt DiGress \citep{vignac2022digress} following the procedure described in Section~\ref{sec:spect/adapting}. We focus on domain-agnostic learning. Our method can be applied as-is to any graph dataset, apart from one-shot variants needing the node frequencies (see Section~\ref{sec:rel_work/graph_gen}). Thus, we use the base version of DiGress, without optimal prior and domain-specific features, replacing them with nodes in-degrees and the number of nodes. As halting and insertion models we use Relational Graph Convolutional Networks \citep{schlichtkrull2018rgcn}. We finetuned the architecture hyperparameters through a Bayesian Search on each dataset. Then, we followed the approach of \cite{huang2023cdgs} and evaluated the sampling quality in several datasets. We use early stopping with validation losses to individually stop each module, as they could have different training times. Time and memory are evaluated using the same hyperparameters to avoid differences in model size. More details on experiments can be found in Appendix~\ref{app:implement}.

\paragraph{Molecular datasets}

We report results on two of the most popular molecular datasets: QM9 \citep{ramakrishnan2014qm9}, and ZINC250k \citep{irwin2012zinc} with 133K and 250K molecules, respectively. As usual, we kekulize the molecules, i.e., remove the hydrogen atoms and replace aromatic bonds with single and double bonds, using the chemistry library RDKit \citep{rdkit}. To measure sample quality we compute the Fréchet ChemNet Distance (FCD) and Neighborhood Subgraph Pairwise Distance Kernel (NSPDK) metrics. We also compute the ratio of Valid, Unique, and Novel molecules, allowing partial charges. For both datasets, we generate 10K molecules, and evaluate FCD and NSPDK on the respective test sets of QM9 and ZINC250k. We use 10\% molecules from QM9 and ZINC250k training sets for validation, respectively.

\paragraph{Generic graphs datasets}
On generic graphs we evaluate our approach on: Community-small, with 100 graphs \citep{you2018graphrnn}; Ego-small and Ego with 200 and 757 graphs \citep{sen2008ego}; Enzymes with 563 protein graphs \citep{schomburg2004enzymes}. We split the train/validation/test sets with the 60/20/20 proportion. We strictly follow \cite{huang2023cdgs} and compute the Maximum Mean Discrepancy (MMD) with radial basis on the distribution of Degree, Clustering coefficient, Laplacian Spectrum coefficient (Spec.) and random GIN embeddings \citep{thompson2022evaluation}, which are a replacement of FCD for generic graphs. For Community-small and Ego-small we generate 1024 graphs, and for Ego and Enzymes we generate the same number of graphs as the test set.

\paragraph{Baselines}
For each dataset, we define 4 degrees of sequentiality of our model: 1-node, small blocks, big blocks, and one-shot. Details on their definition for generic graphs can be found in Table~\ref{tab:generic_degrees}. On molecular datasets we compare versus most of the state-of-the-art models reported in \cite{huang2023cdgs}. Specifically, we consider the autoregressive models GraphAF \citep{shi2020graphaf}, GraphDF \citep{luo2021graphdf}, GraphARM \citep{kong2023grapharm}, and the one-shot models MoFlow \citep{zang2020moflow}, EDP-GNN \citep{niu2020edp_gnn}, GraphEBM \citep{liu2021graphebm}, DiGress \citep{vignac2022digress}, GDSS \citep{jo2022gdss}, CDGS \citep{huang2023cdgs}. On generic graphs datasets, we compare IFH versus the autoregressive models GraphRNN \citep{you2018graphrnn}, GRAN \citep{liao2019gran}, and the one-shot models VGAE \citep{simonovsky2018graphvae}, EDP-GNN \citep{niu2020edp_gnn}, GDSS \citep{jo2022gdss}, CDGS \citep{huang2023cdgs}.

\begin{table}[t]
\centering
\caption{QM9 ablation study for binomial (bin) vs. categorical removal (cat), uniform (unif) vs. BFS ordering.}
\label{tab:qm9_ablation}
\resizebox{\columnwidth}{!}{%
\begin{tabular}{cccccccc}
    \toprule
    Method & Valid$\uparrow$ & Unique$\uparrow$ & Novel$\uparrow$ & NSPDK$\downarrow$ & FCD$\downarrow$ & Time & Memory \\
    & (\%) & (\%) & (\%) & & & (m) & (GB) \\
    \midrule
    bin unif & 91.45 & 97.50 & 94.84 & 6.88e-4 & 1.310 & 61.85 & 1.96 \\
    bin BFS  & 92.72 & 96.34 & 93.01 & 0.001 & 1.175 & 63.00 & 1.96 \\
    cat unif & 89.40 & 98.45 & 93.80 & 4.82e-4 & 1.171 & 28.61 & 0.83 \\
    cat BFS  & 92.30 & 97.70 & 91.81 & 4.78e-4 & 0.918 & 26.48 & 0.80 \\
    \bottomrule
\end{tabular}%
}
\end{table}

\begin{table}[t]
\caption{Sequentiality levels on generic graphs, i.e., block sizes used.}
\label{tab:seq_levels}
\resizebox{\columnwidth}{!}{%
\begin{tabular}{ccccc}
\toprule
    Method & Ego-small & Enzymes & Ego & Comm-small \\
    \midrule
    seq-1 & 1 & 1 & 1 & 1\\
    seq-small & $\{1, 2\}$ & $\{1, 3\}$ & $\{1, 3\}$ & \{1, 2\}\\
    seq-big & $\{1, 2, 8\}$ & $\{1, 2, 8\}$ & $\{1, 4, 16\}$ & \{1, 2, 8\}\\
    one-shot & $n$ & $n$ & $n$ & $n$\\ 
\bottomrule
\end{tabular}%
}
\label{tab:generic_degrees}
\end{table}

\begin{table*}[p]
\caption{Results on the molecule generation task on QM9 (a-c), ZINC250k (d-f) and generic graphs (g) averaged over 3 runs after model selection. For molecular datasets, the tables on the left report performance results, while the tables on the right show the time/memory cost for different levels of sequentiality. On the comparison tables, the best results are in bold, and the second best are underlined.}
\vspace{10pt}
\label{tab:mol_results}
\setcounter{table}{2}
\begin{minipage}[t][0.25\textheight][c]{0.65\textwidth}
\renewcommand{\thefigure}{3a}
\subfloat[Performance results on the QM9 dataset]{
\resizebox{\textwidth}{!}{%
\begin{tabular}{*{7}{c}}
    \toprule
    \multicolumn{2}{c}{Method} & Valid (\%)$\uparrow$ & NSPDK$\downarrow$ & FCD$\downarrow$ & Unique (\%)$\uparrow$ & Novel (\%)$\uparrow$\\
    \midrule
    \multicolumn{2}{c}{Metrics on Training Set} & - & 1.36e-4 & 0.057 & - & -\\
    \midrule
    \multirow{3}{*}{ Autoreg. }
    & GraphAF & 74.43 & 0.021 & 5.625 & 88.64 & 86.59\\
    & GraphDF & 93.88 & 0.064 & 10.928 & 98.58 & 98.54\\
    & GraphARM & 90.25 & 0.002 & 1.220 & 95.62 & 70.39\\
    \midrule
    \multirow{6}{*}{ One-shot }
    & MoFlow & 91.36 & 0.017 & 4.467 & 98.65 & 94.72\\
    & EDP-GNN & 47.52 & 0.005 & 2.680 & 99.25 & 86.58\\
    & GraphEBM & 8.22 & 0.030 & 6.143 & 97.90 & 97.01\\
    & DiGress & 99.00 & 5e-4 & \underline{0.360} & 96.66 & 33.40\\
    & GDSS & 95.72 & 0.003 & 2.900 & 98.46 & 86.27\\
    & CDGS & \underline{99.68} & \underline{3.08e-4} & \textbf{0.200} & 96.83 & 69.62\\
    \midrule
    \multirow{4}{*}{ Ours }
    & seq-1    & \textbf{99.92} & \textbf{2.99e-4} & 0.902 & 96.63 & 88.33\\
    & \{1, 2\} & 94.34 & 4.19e-4 & 0.904 & 97.08 & 89.11\\
    & \{1, 4\} & 92.51 & 7.53e-4 & 0.995 & 97.72 & 92.16\\
    & one-shot & 95.31 & 0.002  & 1.512 & 96.93 & 94.65\\
    \bottomrule
\end{tabular}
}
\label{tab:qm9_results}
}
\end{minipage}
\hfill
\begin{minipage}[t][0.25\textheight][c]{0.31\textwidth}
\subfloat[Training time/memory QM9 dataset]{
\resizebox{\columnwidth}{!}{%
\begin{tabular}{ccc}
    \toprule
    Method & Time/epoch (m) & Memory (GB) \\
    \midrule
    seq-1    & 3.9 & 6.52 \\
    \{1, 2\} & 3.54 & 5.40 \\
    \{1, 4\} & 3.36 & 6.05 \\
    one-shot & 1.98 & 3.73 \\
    \bottomrule
\end{tabular}%
}
\label{tab:qm9_train}
}
\vfill

\subfloat[Generation time/memory QM9 dataset]{
\resizebox{\columnwidth}{!}{%
\begin{tabular}{ccc}
    \toprule
    Method & Time (m) & Memory (GB) \\
    \midrule
    seq-1    & 23.30 & 0.38 \\
    \{1, 2\} & 20.55 & 0.48 \\
    \{1, 4\} & 25.54 & 0.83 \\
    one-shot & 16.92 & 1.22 \\
    \bottomrule
\end{tabular}%
}
\label{tab:qm9_gen}
}
\end{minipage}

\vspace{1cm}

\begin{minipage}[t][0.25\textheight][c]{0.65\textwidth}
\subfloat[Performance results on the ZINC250K dataset]{
\resizebox{\textwidth}{!}{%
\begin{tabular}{*{7}{c}}
    \toprule
    \multicolumn{2}{c}{Method} & Valid (\%)$\uparrow$ & NSPDK$\downarrow$ & FCD$\downarrow$ & Unique (\%)$\uparrow$ & Novel (\%)$\uparrow$\\
    \midrule
    \multicolumn{2}{c}{Metrics on Training Set} & - & 5.91e-5 & 0.985 & - & -\\
    \midrule
    \multirow{3}{*}{ Autoreg. }
    & GraphAF & 68.47 & 0.044 & 16.023 & 98.64 & 99.99\\
    & GraphDF & 90.61 & 0.177 & 33.546 & 99.63 & 100.00\\
    & GraphARM & 88.23 & 0.055 & 16.260 & 99.46 & 100.00\\
    \midrule
    \multirow{6}{*}{ One-shot }
    & MoFlow & 63.11 & 0.046 & 20.931 & 99.99 & 100.00\\
    & EDP-GNN & 82.97 & 0.049 & 16.737 & 99.79 & 100.00\\
    & GraphEBM & 5.29 & 0.212 & 35.471 & 98.79 & 100.00\\
    & DiGress & 91.02 & 0.082 & 23.06 & 81.23 & 100.00\\
    & GDSS & 97.01 & 0.019 & 14.656 & 99.64 & 100.00\\
    & CDGS & \underline{98.13} & \textbf{7.03e-4} & \textbf{2.069} & 99.99 & 99.99\\
    \midrule
    \multirow{4}{*}{ Ours }
    & seq-1 & \textbf{98.56} & \underline{0.002} & \underline{2.387} & 99.87 & 99.89\\
    & \{1, 3\} & 80.59 & 0.004 & 3.312 & 99.98 & 99.95\\
    & \{1, 4, 8\} & 65.68 & 0.015 & 9.229 & 99.94 & 100.00\\
    & one-shot & 60.48 & 0.033 & 15.174 & 100.00 & 100.00\\
    \bottomrule
\end{tabular}%
}
\label{tab:zinc_results}
}
\end{minipage}
\hfill
\begin{minipage}[t][0.25\textheight][c]{0.31\textwidth}
\subfloat[Training time/memory ZINK250K dataset]{
\resizebox{\columnwidth}{!}{%
\begin{tabular}{ccc}
    \toprule
    Method & Time/epoch (m) & Memory (GB) \\
    \midrule
    seq-1    & 30.48 & 15.09 \\
    \{1, 3\} & 20.64 & 16.53 \\
    \{1, 4, 8\} & 20.88 & 15.37 \\
    one-shot & 15.84 & 19.56 \\
    \bottomrule
\end{tabular}%
}
\label{tab:zinc_train}
}

\vfill

\subfloat[Generation time/memory ZINK250K dataset]{
\resizebox{\columnwidth}{!}{%
\begin{tabular}{ccc}
    \toprule
    Method & Time (m) & Memory (GB) \\
    \midrule
    seq-1    & 51.09 & 0.59 \\
    \{1, 3\} & 26.71 & 1.08 \\
    \{1, 4, 8\} & 36.39 & 3.05 \\
    one-shot & 44.43 & 18.03 \\
    \bottomrule
\end{tabular}%
}
\label{tab:zinc_gen}
}
\end{minipage}

\vspace{1cm}

\begin{minipage}[t]{\textwidth}
\subfloat[Performance results on generic graphs datasets]{
\resizebox{\textwidth}{!}{%
\begin{tabular}{*{21}{c}}
    \toprule
    & & \multicolumn{4}{c}{ Community } && \multicolumn{4}{c}{ Ego-small } && \multicolumn{4}{c}{ Enzymes } && \multicolumn{4}{c}{ Ego } \\
    \cline{3-6} \cline{8-11} \cline{13-16} \cline{18-21}
    & & \multicolumn{4}{c}{ $|V|_{\max}=20,\;|E|_{\max}=62$ } && \multicolumn{4}{c}{ $|V|_{\max}=17,\;|E|_{\max}=66$ } && \multicolumn{4}{c}{ $|V|_{\max}=125,\;|E|_{\max}=149$ } && \multicolumn{4}{c}{ $|V|_{\max}=399,\;|E|_{\max}=1071$ }\\
    & & \multicolumn{4}{c}{ $|V|_{\operatorname{avg}}\approx 15,\;|E|_{\operatorname{avg}}\approx 36$ } && \multicolumn{4}{c}{ $|V|_{\operatorname{avg}}\approx 6,\;|E|_{\operatorname{avg}}\approx 9$ } && \multicolumn{4}{c}{ $|V|_{\operatorname{avg}}\approx 33,\;|E|_{\operatorname{avg}}\approx 63$ } && \multicolumn{4}{c}{ $|V|_{\operatorname{avg}}\approx 145,\;|E|_{\operatorname{avg}}\approx 335$ }\\
    \cline{3-6} \cline{8-11} \cline{13-16} \cline{18-21}
    \multicolumn{2}{c}{Method} & Deg.$\downarrow$ & Clus.$\downarrow$ & Spec.$\downarrow$ & GIN$\downarrow$ && Deg.$\downarrow$ & Clus.$\downarrow$ & Spec.$\downarrow$ & GIN$\downarrow$ && Deg.$\downarrow$ & Clus.$\downarrow$ & Spec.$\downarrow$ & GIN$\downarrow$ && Deg.$\downarrow$ & Clus.$\downarrow$ & Spec.$\downarrow$ & GIN$\downarrow$\\
    \midrule
\multicolumn{2}{c}{Metrics on Training Set}  & 0.035 & 0.067 & 0.045 & 0.037 && 0.025 & 0.029 & 0.027 & 0.016 && 0.011 & 0.011 & 0.011 & 0.007 && 0.009 & 0.009 & 0.009 & 0.005 \\
    \midrule
    \multirow{2}{*}{ A-R }
    & GraphRNN & 0.106 & \underline{0.115} & 0.091 & 0.353 && 0.155 & 0.229 & 0.167 & 0.472 && 0.397 & 0.302 & 0.260 & 1.495 && \underline{0.140} & 0.755 & 0.316 & 1.283\\
    & GRAN & 0.125 & 0.164 & 0.111 & 0.196 && 0.096 & 0.072 & 0.095 & 0.106 && 0.215 & 0.147 & 0.034 & 0.069 && 0.594 & \underline{0.425} & 1.025 & 0.244 \\
    \midrule
    \multirow{4}{*}{ O-S }
    & VGAE & 0.391 & 0.257 & 0.095 & 0.360 && 0.146 & 0.046 & 0.249 & 0.089 && 0.811 & 0.514 & 0.153 & 0.716 && 0.873 & 1.210 & 0.935 & 0.520 \\
    & EDP-GNN & \underline{0.100} & 0.140 & 0.085 & \underline{0.125} && \underline{0.026} & \underline{0.032} & 0.037 & \underline{0.031} && 0.120 & 0.644 & 0.070 & 0.119 && 0.553 & 0.605 & 0.374 & 0.295\\
    & GDSS & 0.102 & 0.125 & 0.087 & 0.137 && 0.041 & 0.036 & 0.041 & 0.041 && 0.118 & 0.071 & 0.053 & \underline{0.028} && 0.314 & 0.776 & \underline{0.097} & \underline{0.156}\\
    & CDGS & \textbf{0.052} & \textbf{0.080} & \textbf{0.064} & \textbf{0.062} && \textbf{0.025} & \textbf{0.031} & \underline{0.033} & \textbf{0.025} && \textbf{0.048} & \underline{0.070} & \underline{0.033} & \textbf{0.024} && \textbf{0.036} & \textbf{0.075} & \textbf{0.026} & \textbf{0.026} \\
    \midrule
    \multirow{4}{*}{ Ours }
    & seq-1 & 0.209 & 0.189 & 0.082 & 0.277 && 0.069 & 0.084 & 0.066 & 0.046 && \underline{0.049} & \textbf{0.049} & \textbf{0.026} & 0.088 && 0.303 & 0.643 & 0.311 & 0.352 \\
    & seq-small & 0.177 & 0.167 & 0.082 & 0.203 && 0.031 & 0.041 & 0.040 & 0.043 && 0.252 & 0.237 & 0.077 & 0.404 && 0.435 & 0.898 & 0.162 & 0.403 \\
    & seq-big & 0.141 & 0.173 & 0.089 & 0.262 && 0.027 & 0.042 & \textbf{0.029} &  0.043 && 0.441 & 0.470 & 0.196 & 0.698 && 0.276 & 0.992 & 0.190 & 0.479
 \\
    & oneshot & 0.125 & 0.187 & \underline{0.081} & 0.138 && 0.045 & 0.065 & 0.048 & 0.048 && 0.264 & 0.436 & 0.050 & 0.180 && 0.372 & 0.695 & 0.458 & 0.528 \\
    \bottomrule
\end{tabular}%
}
\label{tab:generic_graphs_results}
}
\end{minipage}
\end{table*}
\subsection{Experimental results}
\label{sec:experim/results}

\paragraph{Ablation study}
We conducted a preliminary ablation study on QM9 (shown in Table~\ref{tab:qm9_ablation}) to evaluate the best-performing formulation for the removal process from those we proposed. For binomial removals, we used the adaptive linear scheduling explained in Appendix~\ref{app:remv_proc/naive/adapt_sched}, and for categorical removals we used $D=\{1, 4\}$ as block sizes, where 4 is approximately half the size of the biggest QM9's molecules. As predicted in Section~\ref{sec:spect/complexity}, the models trained with binomial removals have a huge memory footprint and worse sampling time than categorical removal. When comparing sampling quality, the categorical removal process is still superior. On the ordering, BFS improves quality compared to the uniformly random order, confirming the results of \cite{liao2019gran}.

\paragraph{Performance of the spectrum}
Table~\ref{tab:mol_results} shows that the fully sequential model achieves competitive results with CDGS, surpassing all autoregressive baselines on both QM9 and ZINC250k. Specifically, moving from sequential down shows a drop in general performance. Regarding generic graph generation, we also see competitive results with the state-of-the-art. Still, good performance can also be achieved through small and big block generation, for example, in Ego-small. We observed worse results regarding Community-small and Ego.

\paragraph{Time and memory consumption}
Looking at Tables~\ref{tab:qm9_gen}, \ref{tab:zinc_gen} the level of sequentiality towards 1-node sequential always reduces the memory footprint during generation, as smaller and smaller adjacency matrices are generated, but time goes up, as predicted in Section~\ref{sec:spect/complexity}. Due to the paper's page limit, we refer the reader to Appendix~\ref{app:generics} to find the generic graphs' time/memory consumption tables. We highlight the case in Table~\ref{tab:generic_gen} with the Enzymes and Ego datasets containing very large graphs. On these datasets, the sequential model uses respectively 1/50 and 1/88 of the memory of the one-shot model for generation, although with an increased computational time. During training (Tables~\ref{tab:qm9_train}, \ref{tab:zinc_train}), for small graph datasets such as QM9, memory usage is higher in sequential models, differently from larger graph datasets like ZINC, where the cost of storing big adjacency matrices outweighs that of split sparse graphs.

\section{Discussion}

In Section~\ref{sec:experim}, we showed that adapting our chosen one-shot model to sequential generation led to an improvement of the state-of-the-art for autoregressive generation and being competitive with the state-of-the-art model CDGS. At the same time, one can trade off generation time for memory and performance, although there seems to be a sweet spot inside the spectrum for larger graphs. This shows that the optimal removal process is dataset and task-dependent, and could be considered as a hyperparameter to be tuned when investigating new graph generative models. Our conjecture is that for smaller graph datasets, one-shot models are the fastest and best-performing solution, as seen with the results of CDGS, while as size increases, sequential models should be the go-to, particularly where memory is highly constrained. At huge scales, autoregressive techniques become the only feasible solution \citep{dai2020scalable}. There is still room for improvement on this work's current limitations. For instance, designing better halting models is critical, as larger graphs imply sparser halting signals to train on. Additionally, we found that block-sequential models are susceptible to how information is routed from the previous graph. Then, finding better one-shot models adaptation schemas is crucial.



\section{Conclusion}

In this work, we proposed the IFH framework, which unifies the one-shot and autoregressive paradigms, leaving plenty of room for customization. We showed that high-quality, task-agnostic, autoregressive graph generative models are feasible by adapting DiGress to sequential. In the future, we would like to explore how to better mix the advantages of the two modalities, building upon our framework, gaining the one-shot time efficiency, better memory management, and improved sample quality.



\begin{ack}
We acknowledge the support of the PNRR project FAIR - Future AI Research (PE00000013), under the NRRP MUR program funded by the NextGenerationEU.
\end{ack}


\bibliography{mybibfile}

\newpage

\appendix
\section{Technical Appendix}

\begin{figure*}[t]
\centering
\includegraphics[width=0.75\textwidth]{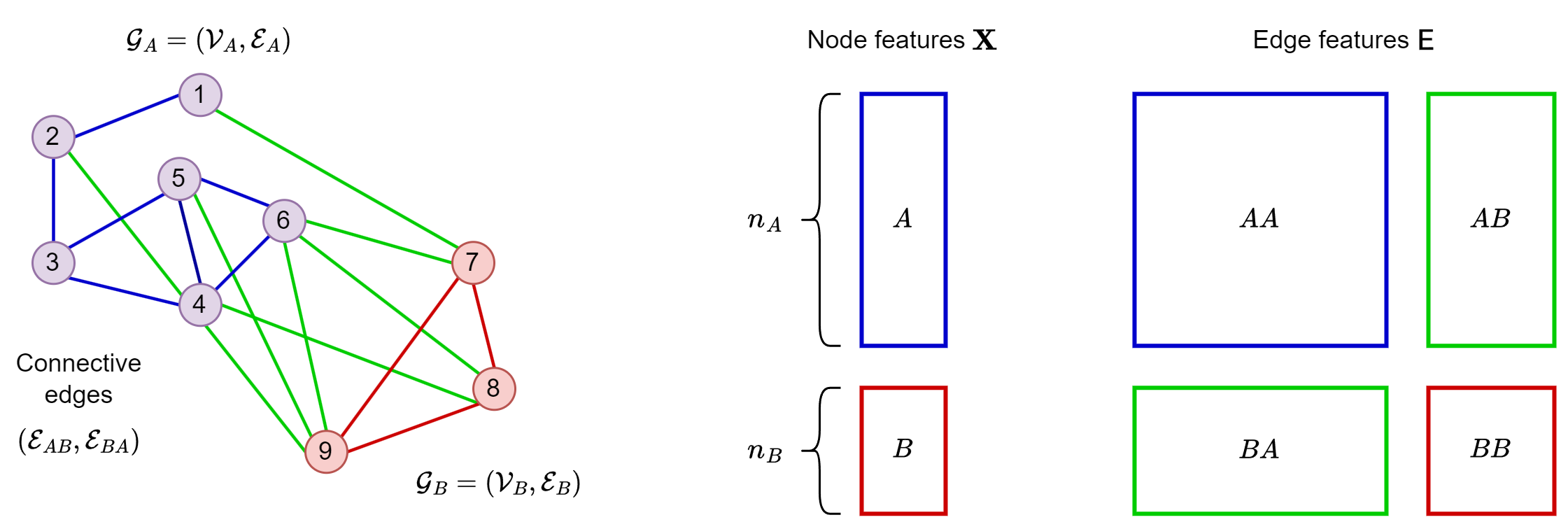}
\caption{Split operation. In blue and red are the induced subgraphs $\gG_A$ and $\gG_B$. In green are the intermediate edges $\mathcal{E}_{AB},\mathcal{E}_{BA}$. On the right is the split adjacency matrix, with the same coloring.}
\label{fig:split_op}
\end{figure*}

\subsection{Detailed definitions}
\label{app:defs}
In this section, we formally define what we colloquially introduced in Section~\ref{sec:background} of the main paper.

\begin{definition}[Remove operation]
\label{def:remv_op}
Removing a node $v_i$ from $\gG$ is equivalent to removing $v_i$ from $\mathcal{V}$, its entry in $\mX$, all edges $(v_i,v_j)$ or $(v_j,v_i)$ from $\mathcal{E}$ in which $v_i$ participates, and the row and column in $\tE$ assigned to its connectivity.
\end{definition}

\begin{definition}[Induced subgraph]
\label{def:induced}
A subgraph $\gG_A$ induced in $\gG$ by $\mathcal{V}_A\subseteq\mathcal{V}$ is the subgraph obtained by removing all nodes in $\mathcal{V}_B=\mathcal{V} \setminus \mathcal{V}_A$ from $\gG$.
\end{definition}

\begin{definition}[Split operation]
\label{def:split_op}
A split $(\gG_A,\gG_B,\mathcal{E}_{AB},\mathcal{E}_{BA})$ of $\gG$ through $\mathcal{V}_A$ is the tuple composed by the subgraphs $\gG_A,\gG_B$ induced by $\mathcal{V}_A$ and $\mathcal{V}_B=\mathcal{V} \setminus \mathcal{V}_A$, the intermediate edges $\mathcal{E}_{AB}$ linking nodes in $\mathcal{V}_A$ to nodes in $\mathcal{V}_B$ and vice versa for $\mathcal{E}_{BA}$.
\end{definition}

\begin{definition}[Merge operation]
\label{def:merge_op}
Given a tuple $(\gG_A,\gG_B,\mathcal{E}_{AB},\mathcal{E}_{BA})$, the merged graph $\gG$ is defined with $\mathcal{V}=\mathcal{V}_A\cup\mathcal{V}_B$ and $\mathcal{E}=\mathcal{E}_A\cup\mathcal{E}_B\cup\mathcal{E}_{AB}\cup\mathcal{E}_{BA}$. Node and edge features are concatenated as shown in Figure~ \ref{fig:split_op}.
\end{definition}

Splitting implies a separation also on features: $\mX_A$ and $\mX_B$ for nodes, $\tE_{AA}$, $\tE_{AB}$, $\tE_{BA}$, and $\tE_{BB}$ for edges, as shown in Figure \ref{fig:split_op}. When splitting undirected graphs, it immediately follows that $\mathcal{E}_{AB}=\mathcal{E}_{BA}$ and $\tE_{AB}=\tE_{BA}^\top$. A merge operation reverses a split operation: in that case, node and edge features are concatenated as shown in Figure~ \ref{fig:split_op}. Now we can define the main object for our mathematical framework, the forward and reversed removal sequences.

\begin{definition}[Forward and reversed removal sequence]
\label{def:remv_seq}
A graph sequence $\gG_{0:T}^{\rightarrow}=(\gG_t)_{t=0}^T$ is a forward removal sequence of $\gG$ when $\gG_0=\gG$, $\gG_T$ is the empty graph $\varnothing$, and $\gG_t$ is an induced subgraph of $\gG_{t-1}$ for all $t=1,\dots,T$. $\gG_{0:T}^{\leftarrow}$ is a reversed removal sequence of $\gG$ if it is a sequence $\gG_{0:T}^{\rightarrow}$ of $\gG$ navigated in reverse, i.e., with index $s=T-t$. In this case $\gG_{s-1}$ is an induced subgraph of $\gG_s$ for all $s=1,\dots,T$.
\end{definition}

We denote $\mathcal{F}(\gG,T)$ and $\mathcal{R}(\gG,T)$ as the sets of all forward and reversed removal sequences of $\gG$ of length $T$. For the halting processes we borrow the notation from \cite{banino2021pondernet}.

\begin{definition}[Halting process]
\label{def:halting}
A halting process $\Lambda_t$ is a Markov process where, at each time step, $\Lambda_t$ is a Bernoulli random variable with outcomes $0,1$ (continue, halt), and evolves as follows: it starts with $\Lambda_0=0$ (continue), and proceeds with Markov transitions $p(\Lambda_t=1|\Lambda_{t-1}=0)=\lambda(t)$ until at step $t=T$ the process is absorbed in state 1 (halt), i.e., $p(\Lambda_t=1|\Lambda_{t-1}=1)=1$ $\forall t>0$.
\end{definition}

\subsection{Generic graphs generation}
\label{app:generics}

\subsubsection{Investigated levels of sequentiality}
In Table \ref{tab:seq_levels} we show our chosen levels of sequentiality, starting from 1-node sequential, then small blocks, then big blocks (also with different sizes), and finally one-shot with $n$ sampled from the dataset empirical distribution on number of nodes. We chose bigger coin denominations for Ego in the seq-big variant, as it contains much larger graphs. Notice that using the categorical removal process (Section \ref{app:remv_proc/cat}), having biggest coin $2$ will roughly reduce the number of steps by two times with respect to 1-node sequential, and so on.

\subsubsection{Detailed discussion on results}

In this section, we expand our findings on generic graphs datasets, which are presented in Table \ref{tab:generic_graphs_results}. Our model is competitive with CDGS in the ego-small and enzymes, but is not on par in Community-small and Ego. We argue the performance in these datasets can be improved by better designing the early stopping mechanism, which might have a positive impact for some datasets, and negatively affect others. Additionally, a better halting mechanism can be helpful for large graphs datasets: particularly for seq-1, the halting signal for training is very sparse. Think of a graph with 500 nodes from Ego, it means that the halting model is trained to predict class 0 (continue) for 499 subgraphs, and class 1 (halt) for the original graph. The same reasoning can be applied to the insertion model, which is trained to use the biggest block size most of the time.

From Table~\ref{tab:generic_gen}, we see that memory usage in generation is always improved by increasing sequentiality, while for training (Table~\ref{tab:generic_train}) it seems to be quite stable. The latter is due to the balancing between the quadratic cost of adjacency matrices, and splitting across steps with smaller block sizes (also discussed in section \ref{sec:spect/complexity}). 

Regarding computational time, we observe that there exist dataset-specific minima. For example, in the Ego dataset with big graphs, seq-big takes the smallest time to run. This might be a sweet spot between how parallel a block generation can be, and the number of steps to generate. The same is observed in Enzymes, where the minimum seems to be between seq-small and seq-big.

\begin{table*}[t]
\caption{Generic graphs results. Note that datasets have different numbers of generated test graphs, so memory and time are not to be compared from one dataset to the other. Training time refers to the time to train for all epochs.}

\subfloat[Training time/memory]{
\resizebox{\textwidth}{!}{%
\begin{tabular}{cccccccccccc}
    \toprule
    & \multicolumn{2}{c}{ Ego-small } && \multicolumn{2}{c}{ Enzymes } && \multicolumn{2}{c}{ Ego } && \multicolumn{2}{c}{ Community-small}\\
    \midrule
    Method & Time/epoch (s) & Memory (GB) &&  Time/epoch (m) & Memory (GB) &&  Time/epoch (m) & Memory (GB) &&  Time/epoch (s) & Memory (GB)\\
    \midrule
    seq-1 & 0.44 & 3.19 && 0.26 & 13.68 && 47.02 & 22.17 && 0.52 & 3.81 \\
    seq-small & 0.33 & 3.22 && 0.18 & 13.15 && 43.05 & 21.97 && 0.39 & 3.21 \\
    seq-big & 0.51 & 7.54 && 0.14 & 14.81 && 16.47 & 22.44 && 0.45 & 4.62 \\
    oneshot & 0.29 & 4.23 && 0.10 & 14.90 && 7.37 & 22.35 && 0.26 & 4.47 \\
    \bottomrule
\end{tabular}%
}
\label{tab:generic_train}
}

\subfloat[Generation time/memory]{
\resizebox{\textwidth}{!}{%
\begin{tabular}{cccccccccccc}
    \toprule
    & \multicolumn{2}{c}{ Ego-small } && \multicolumn{2}{c}{ Enzymes } && \multicolumn{2}{c}{ Ego } && \multicolumn{2}{c}{ Community-small}\\
    \midrule
    Method & Time (m) & Memory (GB) &&  Time (m) & Memory (GB) &&  Time (m) & Memory (GB) &&  Time (m) & Memory (GB)\\
    \midrule
    seq-1 & 4.98 & 0.17 && 31.39 & 0.15 && 458.89 & 0.13 && 7.30 & 0.25\\
    seq-small & 3.36 & 0.20 && 11.36 & 0.19 && 268.89 & 0.17 && 5.62 & 0.30\\
    seq-big & 9.57 &  0.89 && 11.39 & 0.37 && 83.19 & 0.36 && 13.68 & 1.16\\
    oneshot & 5.18 &  1.60 && 23.59 & 7.51 && 202.73 & 11.40 && 7.42 & 2.24\\
    \bottomrule
\end{tabular}%
}
\label{tab:generic_gen}
}
\end{table*}

\subsection{Removal processes}
\label{app:remv_proc}
In this section we provide further details on the removal processes introduced in Section \ref{sec:rem_proc/choose}. All proofs for the equations can be found in Section \ref{app:proofs}.

\subsubsection{Naive (binomial)}
\label{app:remv_proc/naive}
The presented naive method is equivalent to tossing a coin for each node, and removing it for some outcome. A Bernoulli random variable with probability $q_t$ is assigned to each node. All nodes with a positive outcome are removed. The two components of the removal transitions are found to be:
\begin{align}
    q(n_t|\gG_{t-1})=q(n_t|n_{t-1}) &= \binom{n_{t-1}}{n_{t}}q_t^{n_{t-1}-n_t}(1-q_t)^{n_t} \\
    q(\gG_t|n_t,\gG_{t-1}) &= \frac{1}{\binom{n_{t-1}}{n_{t}}}
    \label{eq:bin_trans}
\end{align}
that is, the conditional $n_t|n_{t-1}$ is a Binomial random variable $B(n_t; n_{t-1}, 1-q_t)$, and $\binom{n_{t-1}}{n_{t}}$ are all the ways of choosing $n_t$ nodes from a total of $n_{t-1}$. Furthermore, we can obtain the $t$-step marginal:
\begin{align}
    q(n_t|\gG_{0})&= B(n_t; n_{0}, \pi_t) \\
    q(\gG_t|n_t,\gG_{0}) &= \frac{1}{\binom{n_0}{n_{t}}} \\
    \text{with }\pi_t &= \prod_{k=1}^t(1-q_k) \label{eq:bin_tsteps}
\end{align}
and posterior:
\begin{align}
    q(r_t|\gG_{t},\gG_{0}) &= B(r_t; \Delta n_t, 1-\bar{q}_t) \\
    q(\gG_{t-1}|r_t,\gG_{t},\gG_{0}) &= \frac{1}{\binom{\Delta n_t}{r_t}} \\
    \text{with }\bar{q}_t &=1-\frac{1-\pi _{t-1}}{1-\pi _{t}} \label{eq:bin_post}
\end{align}
where $\Delta n_t=n_{0}-n_{t}$ is the number of removed nodes from step $0$ to step $t$, and as such, can be reinserted to get back $\gG_{t-1}$. The proofs for the equations are found in Section \ref{app:proofs/bin}. Loss \ref{eq:remv_vub} can't be used as it is because there are no reverse distributions for which the KL divergence can be computed without knowing $\Delta n_t$. This is because the support of a Binomial random variable is described by $\Delta n_t$, an information which is not available to the model. For this reason we follow the approach in \cite{austin2021discdiff} and train the insertion model to predict $\Delta n_t$ from $\gG_t$ through an MSE loss, and apply Eq. \ref{eq:bin_post} for sampling.

The hyperparameters $q_t,\pi_t,\bar{q}_t$ can be defined as a schedule on $t$ \citep{ho2020diffusion}. In particular we formulate the schedule in terms of $\pi_t$, which is the average ratio of alive nodes $n_t$ to total nodes $n_0$. We define a linear decay on $\pi_t$:
\begin{equation}
    \pi_t=1-\frac{t}{T}
    \label{eq:linear_decay_sched}
\end{equation}
where $T$ is the number of removal steps as an hyperparameter. At time $t=0$, all nodes are alive ($\pi_0=1$); at time $t=T/2$, half the nodes are alive on average ($\pi_{T/2}=1/2$); at time $t=T$, all nodes have deterministically been removed ($\pi_T=0$). $q_t$ and $\bar{q}_t$ are derived from Equation \ref{eq:linear_decay_sched}:
\begin{align}
    q_t &= 1 - \frac{\pi_t}{\pi_{t-1}}=\frac{1}{T-t+1} \\
    \bar{q}_t &= 1 - \frac{1-\pi_{t-1}}{1-\pi_{t}}=\frac{1}{t}
\end{align}

\subsubsection{Adaptive scheduling}
\label{app:remv_proc/naive/adapt_sched}
With the linear decay schedule, the sizes of blocks depend on the true number of nodes $n_0$, as on average $n_0/T$ nodes are generated. To drop this dependency we make $T$ depend on the number of nodes $n_0$. A way to do so in linear scheduling is by setting:
\begin{equation}
    T = \frac{n_0}{v}, \quad\quad \pi_t = 1-v\frac{t}{n_0}
    \label{eq:adapt_sched}
\end{equation}
where $v$ is the \textit{velocity} hyperparameter. The larger it is, the faster the decay. With this definition, $v$ is also the average number of nodes removed per step, e.g., if a graph has $12$ nodes, and $v=3$, then the graph will become empty in $T=4$ steps, removing on average $3$ nodes at a time. The name velocity comes from the physical interpretation of equation \ref{eq:adapt_sched} as a law of motion.

\subsubsection{Categorical}
\label{app:remv_proc/cat}
The categorical removal process is based on the change-making problem \citep{wright1975change}: let $D\subset\mathbb{N}^d$ denote a set of $d$ coin denominations and, given a total change $C$, we want to find the smallest number of coins needed for making up that amount. This problem can be solved in pseudo-polynomial time using dynamic programming, and knowing the number of coins needed to make up the number of nodes $n_0$ of a graph $\gG_0$ allows to build the shortest possible trajectory $\gG_{0:T}^{\rightarrow}$ using the block size options in $D$. In particular, the number of steps $T$ will always be the number of coins that make the amount $n_0$. To select the number of removed nodes it is enough to pick any permutation of the coins that make $n_0$. This process retains the Markov property because the optimal sequence of coins for $n_t$ is a part of the optimal sequence for $n_0$, if $n_t$ is obtained by any optimal sequence. Categorical transitions describe a distribution on the choices of $D$:
\begin{equation}
    q(r_t|n_{t-1})=\frac{h(n_{t-1})[r_t]}{T-t+1}
    \label{eq:cat_trans}
\end{equation}
where $h(n_{t-1})$ is the histogram on the number of coins in $D$ that make up the amount $n_{t-1}$, $h(n_{t-1})[r_t]$ is the entry corresponding to denomination $r_t$, and $T-t+1$ is the normalization constant, and also the number of coins making up $n_{t-1}$. The $t$-step marginal and posterior distribution can be obtained as:
\begin{align}
    q(n_t|n_0)=\frac{\prod_{d\in D}\binom{h( n_{0})[d]}{h(\Delta n_{t})[d]}}{\binom{T}{t}} \label{eq:cat_tsteps}\\
    q(r_t|n_0,n_t)=\frac{h(\Delta n_t)[r_t]}{t} \label{eq:cat_post}
\end{align}
where $n_t|n_0$ is a multivariate hypergeometric random variable, and $r_t|n_0,n_t$ has the same distribution form of $r_t|n_{t-1}$. The interpretation of the multivariate hypergeometric is that the coins are now colored balls, and an urn contains exactly each of these balls with histogram $h(n_{0})$. We need to shave the amount $\Delta n_t$, so we have to pick exactly the number of balls of each color contained in $h(\Delta n_{t})$. We pick $t$ balls from a total of $T$ in the urn.

\subsubsection{Node ordering}
\label{app:remv_proc/ordering}
Until now we assumed the nodes were removed in a uniformly random order, enforced by the $q(\gG_t|n_t,\gG_{t-1})$, selecting which nodes to keep alive. One example is given by the naive case in Appendix \ref{app:remv_proc/naive}, where nodes are selected uniformly. This doesn't need to be the case, as $q(\gG_t|n_t,\gG_{t-1})$ can actually be any other distribution. Furthermore, to enforce the Markov property once more, we can condition the removal sequence $\gG_{0:T}^{\rightarrow}$ on a particular node ordering $\pi$ before starting the removal. The transitions will then be of the form:
\begin{equation}
    q(\gG_t|\gG_{t-1},\pi)=q(\gG_t|n_t,\gG_{t-1},\pi)q(n_t|\gG_{t-1},\pi)=q(n_t|\gG_{t-1},\pi)
    \label{eq:broken_order}
\end{equation}
The ordering $\pi$ can be taken into account in loss \ref{eq:remv_vub} in the outer expectation. In that case, we have to sample both an example $\gG_0$, and a node ordering $\pi$.

\subsection{Proofs}
\label{app:proofs}

\subsubsection{Proof of the Variational Lower Bound \ref{eq:remv_vub}}
\begin{proof}
Recall the notation in \ref{sec:background} and \ref{app:defs}. To simplify the notation we consider $\mathcal{F}(\gG)$ as the set of any forward removal sequence of $\gG$. Start from the prior distribution of the model:

\begin{align}
p_{\theta ,\phi } &(\gG_{0} ) = \nonumber\\
 & = \sum\limits _{\gG_{1:T}^{\rightarrow} \in \mathcal{F} (\gG_{0})} p_{\theta ,\phi } (\gG_{0:T}^{\rightarrow}) \\
 & =\sum\limits _{\gG_{1:T}^{\rightarrow} \in \mathcal{F} (\gG_{0})} p_{\theta ,\phi } (\gG_{0:T}^{\rightarrow})\frac{q(\gG_{1:T}^{\rightarrow} |\gG_{0} )}{q(\gG_{1:T}^{\rightarrow} |\gG_{0} )} \label{proof:imp_sampl}\\
 & =\sum\limits _{\gG_{1:T}^{\rightarrow} \in \mathcal{F} (\gG_{0})} q(\gG_{1:T}^{\rightarrow} |\gG_{0} )p_{\theta} (\gG_{T} )\frac{p_{\theta ,\phi } (\gG_{0:T-1} |\gG_{T} )}{q(\gG_{1:T}^{\rightarrow} |\gG_{0} )}\\
 & =\sum\limits _{\gG_{1:T}^{\rightarrow} \in \mathcal{F} (\gG_{0})} q(\gG_{1:T}^{\rightarrow} |\gG_{0} )p_{\theta} (\gG_{T} )\prod _{t=1}^{T}\frac{p_{\theta ,\phi } (\gG_{t-1} |\gG_{t} )}{q(\gG_{t} |\gG_{t-1})} \label{proof:markov}\\
 & =\sum\limits _{\gG_{1:T}^{\rightarrow} \in \mathcal{F} (\gG_{0})} q(\gG_{1:T}^{\rightarrow} |\gG_{0} )\frac{p_{\theta} (\gG_{T} )}{q( \gG_{T} |\gG_{0})} p_{\theta ,\phi } (\gG_{0} |\gG_{1} )\nonumber\\
 &\quad\quad\quad\cdot \prod _{t=2}^{T}\frac{p_{\theta ,\phi } (\gG_{t-1} |\gG_{t} )}{q(\gG_{t-1} |\gG_{t} ,\gG_{0})} \\
 & =\sum\limits _{\gG_{1:T}^{\rightarrow} \in \mathcal{F} (\gG_{0})} q(\gG_{1:T}^{\rightarrow} |\gG_{0} )\frac{p_{\theta} (\gG_{T} )}{q(\gG_{T} |\gG_{0})} p_\phi (n_{0} |\gG_{1} ) p_{\theta} (\gG_{0} |n_{0} ,\gG_{1} ) \nonumber\\
 &\quad\quad\quad\cdot \prod _{t=2}^{T}\frac{p_\phi (n_{t-1} |\gG_{t} )}{q( n_{t-1} |\gG_{t} ,\gG_{0})}\frac{p_{\theta} (\gG_{t-1} |n_{t-1} ,\gG_{t} )}{q(\gG_{t-1} |n_{t-1} ,\gG_{t} ,\gG_{0})}\\
 & =\sum\limits _{\gG_{1:T}^{\rightarrow} \in \mathcal{F} (\gG_{0})} q(\gG_{1:T}^{\rightarrow} |\gG_{0} )\frac{p_{\theta} (\gG_{T} )}{q(\gG_{T} |\gG_{0})} p_\phi (r_1 |\gG_{1} ) p_{\theta} (\gW_1 |r_1 ,G_{1} )\nonumber\\
 &\quad\quad\quad\cdot \prod _{t=2}^{T}\frac{p_\phi (r_t |\gG_{t} )}{q( r_t |\gG_{t} ,\gG_{0})}\frac{p_{\theta} (\gW_t |r_t ,\gG_{t} )}{q( \gW_t |r_t ,\gG_{t} ,\gG_{0})}\\
 & =\sum\limits _{\gG_{1:T}^{\rightarrow} \in \mathcal{F} (\gG_{0})} q(\gG_{1:T}^{\rightarrow} |\gG_{0} ) p_\phi (r_1 |\gG_{1} ) p_{\theta} (\gW_1 |r_1 ,G_{1} )\nonumber\\
 &\quad\quad\quad\cdot \prod _{t=2}^{T}\frac{p_\phi (r_t |\gG_{t} )}{q( r_t |\gG_{t} ,\gG_{0})}\frac{p_{\theta} (\gW_t |r_t ,\gG_{t} )}{q( \gW_t |r_t ,\gG_{t} ,\gG_{0})}
\end{align}
Some significant steps are \ref{proof:imp_sampl}, where we used importance sampling, \ref{proof:markov} where we factorized the probabilities over sequences with their definitions

The Variational Upper Bound is found from the negative log likelihood through the Jensen Inequality:
\begin{align*}
    \E_{q(\gG_{0})}&[-\log p_{\theta ,\phi } (\gG_{0} )] \leq 
    \E_{q(\gG_{0})}\Bigg[\sum\limits_{t=2}^{T}\KL \big(q(r_t|\gG_t,\gG_0) \Vert p_{\phi}(r_t|\gG_t)\big)+ \nonumber\\ 
    &-\E_{q(\gG_1|\gG_0)}\left[\log p_\phi(r_{1}|\gG_1)\right]+ \nonumber\\
	&+\sum\limits_{t=2}^{T}\KL \big(q(\gW_t|r_t,\gG_t,\gG_0) \Vert p_{\theta}(\gW_t|r_t,\gG_t)\big)+ \nonumber\\
 & -\E_{q(\gG_1|\gG_0)}\left[\log p_\theta(\gW_1|r_1,\gG_1)\right]
	\Bigg]
\end{align*}
\end{proof}

\subsubsection{Binomial removal}
\label{app:proofs/bin}

\paragraph{Proof of equation \ref{eq:bin_tsteps}}
\begin{proof}
Let's prove by induction. Consider the simple case for $n_{1}$:
$$
q(n_{1}|n_{0})=B(n_{1};n_0,\pi_{1})
$$
with $\pi_{1}=1-q_{1}$. This is true by the definition of binomial transitions \ref{eq:bin_trans}.

Now, assume the property is true for $t-1$, that is, $n_{t-1}|n_{0}$ is a Binomial $B(n_{t-1};n_0,\pi_{t-1})$. We know that $n_{t}|n_{t-1}$ is also a Binomial, and has the same distribution as $n_{t}|n_{t-1},n_{0}$ due to the Markov property. Let's recall what their distribution and parameters are:
\begin{align*}
    n_{t}|n_{t-1},n_{0}&\sim B(n_{t}; n_{t-1}|n_{0},1-q_{t}) \\
n_{t-1}|n_{0}&\sim B(n_{t-1}; n_{0},\pi_{t-1})\quad\quad \pi_{t-1}=\prod\limits_{k=1}^{t-1}(1-q_{k})
\end{align*}

It can be proven that a Binomial conditioned on a Binomial is still a Binomial with success probability the product of the two success probabilities, and number of experiments the same as the conditioning binomial. From this fact $n_{t}|n_{0}$ is a Binomial:
\begin{equation*}
    n_{t}|n_{0}\sim B(n_t; n_{0},\pi_{t})\quad\quad \pi_{t}=(1-q_{t})\pi_{t-1}=\prod\limits_{k=1}^{t}(1-q_{k})
\end{equation*}
\end{proof}

\paragraph{Proof of equation \ref{eq:bin_post}}

\begin{proof}
Let's compute the posterior:
\begin{align*}
q(n_{t-1}&|n_{t} ,n_{0} ) = \\
 & = q(n_{t} |n_{t-1} )\frac{q(n_{t-1} |n_{0} )}{q(n_{t} |n_{0} )}\\
 & =\frac{n_{t-1} !}{n_{t} !(n_{t-1} -n_{t} )!} (1-q_{t})^{n_{t}} q_{t}^{n_{t-1} -n_{t}} \\
 &\quad\quad\quad\cdot \frac{\frac{n_{0} !}{n_{t-1} !(n_{0} -n_{t-1} )!} \pi _{t-1}^{n_{t-1}} (1-\pi _{t-1} )^{n_{0} -n_{t-1}}}{\frac{n_{0} !}{n_{t} !(n_{0} -n_{t} )!} \pi _{t}^{n_{t}} (1-\pi _{t} )^{n_{0} -n_{t}}}\\
 & =\frac{(n_{0} -n_{t} )!}{(n_{t-1} -n_{t} )!(n_{0} -n_{t-1} )!} \pi _{t-1}^{n_{t-1} -n_{t}} q_{t}^{n_{t-1} -n_{t}}\\
 &\quad\quad\quad\cdot \frac{(1-\pi _{t-1} )^{n_{0} -n_{t-1}}}{(1-\pi _{t} )^{n_{0} -n_{t}}}\\
 & =\frac{(n_{0} -n_{t} )!}{(n_{t-1} -n_{t} )!(n_{0} -n_{t} -( n_{t-1} -n_{t}) )!} \\
 &\quad\quad\quad\cdot \pi _{t-1}^{n_{t-1} -n_{t}} (1-\pi _{t-1} )^{n_{0} -n_{t-1}}\frac{q_{t}^{n_{t-1} -n_{t}}}{(1-\pi _{t} )^{n_{0} -n_{t}}}\\
 & =\binom{n_{0} -n_{t}}{n_{t-1} -n_{t}}\left( q_{t}\frac{\pi _{t-1}}{1-\pi _{t}}\right)^{n_{t-1} -n_{t}}\left( q_{t}\frac{1-\pi _{t-1}}{1-\pi _{t}}\right)^{n_{0} -n_{t-1}}\\
 &=\binom{n_{0} -n_{t}}{n_{0} -n_{t-1}}\left(\frac{1-\pi _{t-1}}{1-\pi _{t}}\right)^{n_{0} -n_{t-1}}\left( 1-\frac{1-\pi _{t-1}}{1-\pi _{t}}\right)^{n_{t-1} -n_{t}}
\end{align*}

Finally, by substituting the number of failures at step $t$: $r_{t}=n_{t-1}+n_{t}$ we get:
$$
 q(r_{t} |n_{t} ,n_{0} ) =\binom{n_{0} -n_{t}}{r_{t}}\left( q_{t}\frac{\pi _{t-1}}{1-\pi _{t}}\right)^{r_{t}}\left(\frac{1-\pi _{t-1}}{1-\pi _{t}}\right)^{n_{0} -n_{t} -r_{t}}
 $$
 \end{proof}

\subsubsection{Categorical removal}
\label{app:proofs/cat}

\paragraph{Proof of equation \ref{eq:cat_tsteps}}

\begin{proof}
Let's prove this by induction. Consider the simple case for $n_{1}$:
$$
q(n_{1}|n_{0})=q(r_1|n_{0})
=\frac{\prod_{d\in D}\binom{h( n_{0})[d]}{h(r_1)[d]}}{\binom{T}{1}}
=\frac{h( n_{0})[r_1]}{T}
$$
where the product over denominations only one non-unit factor with $d=r_1$, because $h(r_1)[r_1]=1$ and $h(r_1)[d]=0$ for all other denominations, as $r_1$ is one of the possible choices in $D$.

Now, assume the property is true for $t-1$, that is, $n_{t-1}|n_{0}$ is a Multivariate hypergeometric, that is:

\begin{equation}
q(n_{t-1}|n_0)=\frac{\prod_{d\in D}\binom{h( n_{0})[d]}{h(\Delta n_{t-1})[d]}}{\binom{T}{t-1}}
\end{equation}

Now, using the law of total probability:
\vspace{-3pt}

\begin{align*}
q(n_t|n_0)
&= \sum_{n_{t-1}=n_t}^{n_0}q(n_t|n_{t-1})q(n_{t-1}|n_0) \\
&= \sum_{d\in D} q(n_t|n_t+d)q(n_t+d|n_0) \\
&= \sum_{d\in D} \frac{h(n_t+d)[d]}{T-t+1}
\frac{\prod_{d'\in D}\binom{h( n_{0})[d']}{h(n_0 - n_t - d)[d']}}{\binom{T}{t-1}} \\
&=\frac{1}{(T-t+1)\frac{T!}{(T-t+1)!(t-1)!}}\sum _{d\in D} h(n_{t} +d)[d]\\
&\quad\quad\quad\cdot \prod _{d'\in D}\binom{h(n_{0} )[d']}{h(n_{0} -n_{t} -d)[d']} \\
&=\frac{1}{t}\frac{1}{\frac{T!}{(T-t)!t!}}\sum _{d\in D}( h(n_{t} )[d]+1)\\
&\quad\quad\quad\cdot \frac{h(n_{0} )[d]!}{( h(n_{0} )[d]-h(n_{0} -n_{t} -d)[d]) !h(n_{0} -n_{t} -d)[d]!} \\
&\quad\quad\quad\cdot \prod _{d'\in D\setminus \{d\}}\binom{h(n_{0} )[d']}{h(n_{0} -n_{t} -d)[d']}\\
&=\frac{1}{t}\frac{1}{\binom{T}{t}}\sum _{d\in D}( h(n_{t} )[d]+1)\\
&\quad\quad\quad\cdot \frac{h(n_{0} )[d]!}{( h(n_{0} )[d]-h(n_{0} -n_{t} )[d]+1) !( h(n_{0} -n_{t} )[d]-1) !}\\
&\quad\quad\quad\cdot \prod _{d'\in D\setminus \{d\}}\binom{h(n_{0} )[d']}{h(n_{0} -n_{t} )[d']}\\
&=\frac{1}{t}\frac{1}{\binom{T}{t}}\sum _{d\in D}( h(n_{t} )[d]+1)\\
&\quad\quad\quad\cdot \frac{h(n_{0} )[d]!}{( h(n_{t} )[d]+1) !( h(n_{0} )[d]-h(n_{t} )[d]-1) !} \\
&\quad\quad\quad\cdot \prod _{d'\in D\setminus \{d\}}\binom{h(n_{0} )[d']}{h(n_{0} -n_{t} )[d']}\\
&=\frac{1}{t}\frac{1}{\binom{T}{t}}\sum _{d\in D} h(n_{0} -n_{t} )[d]\frac{h(n_{0} )[d]!}{h(n_{t} )[d]!( h(n_{0} )[d]-h(n_{t} )[d]) !}\\
&\quad\quad\quad\cdot \prod _{d'\in D\setminus \{d\}}\binom{h(n_{0} )[d']}{h(n_{0} -n_{t} )[d']}\\
\end{align*}
\begin{align*}
&=\frac{1}{t}\frac{1}{\binom{T}{t}}\sum _{d\in D} h(n_{0} -n_{t} )[d]\binom{h(n_{0} )[d]}{h(n_{0} -n_{t} )[d]}\\
&\quad\quad\quad\cdot \prod _{d'\in D\setminus \{d\}}\binom{h(n_{0} )[d']}{h(n_{0} -n_{t} )[d']}\\
&=\frac{1}{t}\frac{1}{\binom{T}{t}}\sum _{d\in D} h(n_{0} -n_{t} )[d]\prod _{d'\in D}\binom{h(n_{0} )[d']}{h(n_{0} -n_{t} )[d']}\\
&=\frac{1}{t}\frac{\prod _{d'\in D}\binom{h(n_{0} )[d']}{h(n_{0} -n_{t} )[d']}}{\binom{T}{t}}\sum _{d\in D} h(n_{0} -n_{t} )[d]\\
&=\frac{1}{t}\frac{\prod _{d'\in D}\binom{h(n_{0} )[d']}{h(n_{0} -n_{t} )[d']}}{\binom{T}{t}} t\\
&=\frac{\prod _{d'\in D}\binom{h(n_{0} )[d']}{h(\Delta n_{t} )[d']}}{\binom{T}{t}}
\end{align*}

To reach the final statement we used the following facts:
\begin{itemize}
    \item $h(n+d)[d']=h(n)[d']$ for all components $d'\neq d$, and $h(n+d)[d]=h(n)[d]+1$
    \item $h(n_0)-h(n_t)=h(n_0-n_t)$
    \item by definition $\sum_{d\in D}h(n_0-n_t)[d]=t$
\end{itemize}
\end{proof}

\paragraph{Proof of equation \ref{eq:cat_post}}

\begin{proof}
Let's compute the posterior:
\begin{align*}
q( n_{t-1} |n_{0} ,n_{t}) &= \frac{q( n_{t} |n_{t-1}) qp( n_{t-1} |n_{0})}{q( n_{t} |n_{0})} \\
&=\frac{h( n_{t-1})[ r_{t}]}{T-t+1}\frac{\prod _{d\in D}\frac{h( n_{0})[ d] !}{h( \Delta n_{t-1})[ d] ! h( n_{t-1})[ d!}}{\frac{T!}{( t-1) !( T-t+1) !}}\\
&\quad\quad\quad\cdot \left(\frac{\prod _{d\in D}\frac{h( n_{0})[ d] !}{h( \Delta n_{t})[ d] ! h( n_{t})[ d] !}}{\frac{T!}{t!( T-t) !}}\right)^{-1}\\
&=\frac{h( n_{t-1})[ r_{t}]( t-1) !( T-t+1) !}{t!( T-t) !( T-t+1)}\\
&\quad\quad\quad\cdot \prod _{d\in D}\frac{h( \Delta n_{t})[ d] ! h( n_{t})[ d] !}{h( \Delta n_{t-1})[ d] ! h( n_{t-1})[ d] !}\\
&=\frac{h( n_{t-1})[ r_{t}]}{t}\prod _{d\in D}\frac{h( \Delta n_{t})[ d] ! h( n_{t})[ d] !}{h( \Delta n_{t-1})[ d] ! h( n_{t-1})[ d] !}\\
&=\frac{h( n_{t-1})[ r_{t}]}{t}\prod _{d\in D}\frac{h( \Delta n_{t-1} +r_{t})[ d] ! h( n_{t})[ d] !}{h( \Delta n_{t-1})[ d] ! h( n_{t} +r_{t})[ d] !}\\
&=\frac{h( n_{t-1})[ r_{t}]}{t}\frac{h( \Delta n_{t-1} +r_{t})[ r_{t}] ! h( n_{t})[ r_{t}] !}{h( \Delta n_{t-1})[ r_{t}] ! h( n_{t} +r_{t})[ r_{t}] !}\\
&\quad\quad\quad\cdot \prod _{d\in D\setminus \{r_{t}\}}\frac{h( \Delta n_{t-1})[ d] ! h( n_{t})[ d] !}{h( \Delta n_{t-1})[ d] ! h( n_{t})[ d] !}\\
&=\frac{h( n_{t-1})[ r_{t}]}{t}\frac{( h( \Delta n_{t-1})[ r_{t}] +1) ! h( n_{t})[ r_{t}] !}{h( \Delta n_{t-1})[ r_{t}] ! ( h( n_{t})[ r_{t}] +1) !}\\
&=\frac{h( n_{t})[ r_{t}] +1}{t}\frac{h( \Delta n_{t-1})[ r_{t}] +1}{h( n_{t})[ r_{t}] +1}\\
&=\frac{h( \Delta n_{t})[ r_{t}]}{t}
\end{align*}

Because $q(n_{t-1}|n_0,n_t)=q(r_t|n_0,n_t)$:
$$
q(r_t|n_0,n_t)=\frac{h( \Delta n_{t})[ r_{t}]}{t}
$$

\end{proof}

\subsection{Implementation details}
\label{app:implement}

We implemented our framework using PyTorch \citep{paszke2019pytorch}, PyTorch Lightning \citep{Falcon_PyTorch_Lightning_2019} and PyTorch Geometric \citep{torch_geom}. Our foundation was the DiGress implementation \citep{vignac2022digress}, which we heavily modified and partly reimplemented to generalize on many cases. The source code can be found at \url{https://github.com/CognacS/ifh-model-graphgen}.

We run a Bayesian Hyperaparameter Search for each dataset-sequentiality degree pair, with the only exception of ZINC250k, which is the most computationally intensive dataset due to its size. We validated on 15 runs for each pair and picked the hyperparameters which yielded the best validation loss values. For assessing halting, we computed the Earth-Mover distance with respect to the prior distribution of having halted at each step, which we found to capture well the quality of halting. We then adopted these hyperparameters for our final experiments. For ZINC250k we adopted a set of hyperparameters which we found successful, taking inspiration from those given by DiGress.

All our search procedure parameters, experiments, and their hyperparameters are available in our code as simple Hydra \citep{Yadan2019Hydra} configuration files. Each was run for 3 different seeds. For each experiment, we also report the time to sample the set of generated graphs and the memory footprint. We ran ZINC250k experiments on a V100 GPU, Ego experiments on an L4 GPU, and all other experiments on a T4 GPU.

We implemented the insertion model and halting model (when needed) as RGCN \citep{schlichtkrull2018rgcn} to tackle labelled datasets, and GraphConvs \citep{morris2019graphconv} for unlabelled datasets. We implemented the halting model in the same way.

\subsubsection{Adapting DiGress}

We briefly discuss how we adapted the DiGress model and architecture to act as a filler model. The nodes of the already generated graph are encoded through an RGCN or GraphConv, and are used as input in the graph transformer architecture \citep{dwivedi2020graph_transf}, together with the vectors of noisy labels of the new nodes. Noisy edges are sampled both between new nodes, and also from new nodes to existing nodes. In a graph transformer layer, new nodes can attend both to themselves and old nodes, and mix with the information on edges, as is done in DiGress. Finally, the vectors of new nodes and edges are updated through the Feed Forward Networks of the transformer layer, while the encoded old nodes remain untouched. With this last consideration, one can encode the nodes of the already generated graph only once in a filler model call, and use them in all the DiGress denoising steps.

\end{document}